\documentclass[review]{elsarticle}

\usepackage{lineno,hyperref}
\usepackage{subfig}
\usepackage{graphicx}
\usepackage[table,xcdraw]{xcolor}
\usepackage{amsmath,amssymb}
\usepackage{amsmath}
\modulolinenumbers[5]

\journal{Neural Networks}

\begin{document}

\begin{frontmatter}

\title{Neural Memory Plasticity for Anomaly Detection}

\author[label1]{Tharindu Fernando\corref{cor1}}
\ead{t.warnakulasuriya@qut.edu.au}
 
\author[label1]{Simon~Denman}
\ead{s.denman@qut.edu.au}
 
\author[label1]{David Ahmedt-Aristizabal}
\ead{david.aristizabal@hdr.qut.edu.au}

\author[label1]{Sridha Sridharan}
\ead{s.sridharan@qut.edu.au}

\author[label2]{Kristin Laurens}
\ead{kristin.laurens@qut.edu.au}
 
\author[label2]{Patrick Johnston}
\ead{patrick.johnston@qut.edu.au}
 
\author[label1]{ Clinton~Fookes}
\ead{c.fookes@qut.edu.au}

\cortext[cor1]{Corresponding author at: Image and Video Research Laboratory, SAIVT, Queensland University of Technology, Australia.}
\address[label1]{Image and Video Research Laboratory, SAIVT, Queensland University of Technology, Australia.}
\address[label2]{School of Psychology and Counselling and the Institute of Health Biomedical Innovation (IHBI), Queensland University of Technology, Australia.}

\begin{abstract}
In the domain of machine learning, Neural Memory Networks (NMNs) have recently achieved impressive results in a variety of application areas including visual question answering, trajectory prediction, object tracking, and language modelling. However, we observe that the attention based knowledge retrieval mechanisms used in current NMNs restricts them from achieving their full potential as the attention process retrieves information based on a set of static connection weights. This is suboptimal in a setting where there are vast differences among samples in the data domain; such as anomaly detection where there is no consistent criteria for what constitutes an anomaly. In this paper, we propose a plastic neural memory access mechanism which exploits both static and dynamic connection weights in the memory read, write and output generation procedures. We demonstrate the effectiveness and flexibility of the proposed memory model in three challenging anomaly detection tasks in the medical domain: abnormal EEG identification, MRI tumour type classification and schizophrenia risk detection in children. In all settings, the proposed approach outperforms the current state-of-the-art. Furthermore, we perform an in-depth analysis demonstrating the utility of neural plasticity for the knowledge retrieval process and provide evidence on how the proposed memory model generates sparse yet informative memory outputs. 
\end{abstract}


\end{frontmatter}


\section{Introduction}

Neural Memory Networks (NMNs) have recently achieved tremendous success on larger knowledge bases via the use of an external memory to explicitly store and retrieve relevant information \cite{munkhdalai2017neural,kumar2016ask,xiong2016dynamic,fernando2018task,yang2018learning,fernando2018tree}. They elevate the temporal model capability of Recurrent Neural Networks (RNNs) \cite{hochreiter2001gradient} to capture both long-term and short-term relationships where the added capacity is utilised to store the information that constitutes these relationships. The evolution of the memory occurs through read and write functions, which are both differentiable and trained alongside the rest of the components of the network.  

Plasticity is a biological process which refers to the human brain's ability to change throughout life by forming new connections among neurons and degrading unwanted connections \cite{perwej2012neuroplasticity}. Most recently, in \cite{miconi2018differentiable} the authors employ plasticity in neural networks and optimise it along with the rest of the parameters through back propagation. Their evaluations demonstrate encouraging results with the addition of dynamicity in connections to capture temporal relationships. However, we argue that the full potential of plasticity in knowledge discovery tasks is yet to be explored as plasticity has only been exploited for vanilla neural networks and they have not ventured into memory networks. Hence, their capacity for knowledge discovery is yet to be fully enabled. 

Although plasticity in neural networks has some potential to model temporal relationships, their capacity is limited, hence, they fail to recover long-term dependencies. On the other hand, even though a memory network is analogous to that of the human brain, the naive structure of the read and write mechanisms in NMNs hinder their potential for knowledge discovery. Firstly, the attention mechanism employed for the extraction of relevant information tries to embed the stored knowledge as a fixed dimensional vector. In contrast, in biological brains, the connections are changing with the aid of plasticity \cite{miconi2018differentiable}. Secondly, the Long Short-Term Memory (LSTM) function which is commonly employed in NMN to read from and write to the memory over time is proven to focus more on recent history as opposed to discovering long-term relationships \cite{chen2016enhancing}. Via carefully evaluating the merits of both neural plasticity and NMN techniques we propose a new knowledge retrieval structure to be used by NMNs through trainable neural plasticity. In order to demonstrate the ability of our proposed architecture in achieving the full potential of neural plasticity we evaluate our memory model in an anomaly detection setting.  Anomaly detection is a fundamental, yet challenging task in machine learning, primarily due to the lack of a consistent criteria for what constitutes an anomaly. A possible solution to this problem is to develop a memory architecture which would optimally compare and contrast between different characteristics that arise through analysing the long-term relationships within the data domain.  

A plastic memory network architecture as we have proposed would allow the underlying framework to learn a vast range of subject and problem specific characteristics from the data via temporally varying the level of attention that it pays in the memory read and write operations to different salient information cues. As a result, the same underlying approach can be applied to a range of applications. To demonstrate this ability, we explore a range of anomaly detection tasks in the medical domain. The application of machine learning techniques to automatically detect anomalies in medical data is particularly attractive considering its consistency and non-subjectivity, along with its cost-effectiveness, eliminating the need for extensive training of human practitioners which is required to master manual screening \cite{mahmud2018applications}. There exists numerous medical anomaly detection tasks, ranging from identifying abnormal EEG recordings \cite{yildirim2018deep}, detecting tumours in Magnetic Resonance Imaging (MRI) \cite{afshar2018brain} to anomaly detection in medical wireless sensor networks \cite{pachauri2015anomaly}. However, medical data itself poses new challenges as there exists significant variability among subjects, and across different conditions. For instance, identifying an anomaly in an Electroencephalogram (EEG) is inherently difficult even for trained professionals, as there exists significant variability among the patients in the manifestation of any abnormality which is accentuated further by the variability in the operating conditions \cite{yildirim2018deep}. 

To demonstrate the applicability of our proposed technique to a breadth of applications in the medical domain, this paper investigates abnormal EEG identification, MRI tumour type classification, and schizophrenia risk detection tasks, and proposes a unified system which learns these patient-specific and problem-specific characteristics from the data. Our evaluations on these challenging abnormality detection tasks, which involves both one and two-dimensional signals, confirms the viability of the method for real-world applications. Through extensive experimental evaluations, we demonstrate that neural plasticity enhances the knowledge retrieval process in NMNs where the memory is translated into very different forms, which are learned over time, and allows us to filter out the most salient information. Furthermore, we analyse the utility of plasticity in terms of model activations and statistical interactions, and demonstrate how it acts as an attention mechanism during memory access. The main contributions of the proposed work can be summarised as follows:

\begin{itemize}
\item We propose a novel memory addressing mechanism for NMNs which facilitates trainable neural plasticity, allowing better extraction of stored knowledge. 
\item We outperform state-of-the-art methods in three challenging tasks in the medical domain: abnormality detection in EEGs, detecting schizophrenia risks in children using EEG recordings, and classification of tumour types in MRIs. These applications involve both one and two dimensional signals, and binary and multi-class classification tasks. 
\item We interpret the model learning process in terms of activations from the memory model and statistical interactions \cite{tsang2017detecting} and illustrate the importance of neural plasticity for salient information retrieval. 
\end{itemize}

We would like to emphasise that even though we are demonstrating our approach on three different applications specifically in the medical domain, the varied nature of these problems demonstrates how the proposed model can be directly applied to any anomaly detection problem in different domains where modelling long term relationships is necessary. Possible application areas include, detecting anomalies in daily human activities and sports activities \cite{ghafoori2018efficient}, anomaly detection in vehicle driving \cite{zhang2017safedrive}, and detecting anomalies in stock exchange \cite{cao2014adaptive} and in credit card transactions \cite{dal2017credit}. 

\section{Related Work}

\subsection{Neural Memory Networks} 
The authors in \cite{munkhdalai2017neural,yang2018learning,xiong2016dynamic,kumar2016ask} and our prior works \cite{fernando2018tree,fernando2018pedestrian,fernando2018learning,fernando2018task} have extensively demonstrated the effectiveness of what are termed  ``memory modules'' to store and retrieve relevant information, and capture relationships between different input sequences in the data domain. These dependencies are missed by models such as LSTMs and Gated Recurrent Units (GRUs) as they consider dependencies only within a given input sequence. Due to this ability external memory modules have gained traction in numerous domains, including language modelling \cite{munkhdalai2017neural}, visual question answering \cite{xiong2016dynamic}, trajectory prediction \cite{fernando2018tree,fernando2018pedestrian}, object tracking \cite{yang2018learning}, and saliency modelling \cite{fernando2018task}. 

Even through they exhibit great potential for capturing salient information, we observe several factors that hinder their capabilities. Firstly, the LSTM functions utilised for the memory read and write processes are demonstrated to focus more on the recent dependencies and completely ignore long-term dependencies \cite{chen2016enhancing}. Furthermore, recent works \cite{miconi2018differentiable} have shown that the attention based knowledge retrieval mechanism is not ideal in a memory unit as the stored information is temporally evolving, requiring the weighted connections to be updated over time. 

Inspired by the works of Miconi et. al \cite{miconi2018differentiable} where they introduce the concept of differentiable plasticity, we propose a novel mechanism to retrieve and update neural memory models. It should be noted that \cite{miconi2018differentiable} investigated the plasticity only in vanilla deep neural networks such as LSTMs, hence its full potential for knowledge discovery is yet to be fully exploited. 

\subsection{Neural Plasticity}
``Neural plasticity'', the strengthening and weakening of connections between the neurons using the neural activity as the basis, has been extensively investigated within artificial neural networks \cite{soltoggio2018born}. However, these investigations were conducted before the dawn of deep learning, hence, requiring extensive research to utilise its full potential in knowledge retrieval. Plastic methods build upon the ``Hebbian rule'', neurons that fire together, wire together \cite{hebb2005organization}. For instance in \cite{nolfi1993auto} the authors Nolfi and Parisi propose to evolve networks with ``auto-teaching'' inputs and utilise them to provide an error signal for the network weight adjustment over their lifetime. In \cite{rolls2000design} the authors propose eight different rules to inject the biological evolution of rules when updating the neural network parameters. In \cite{maniadakis2006modelling} the authors use separate neural networks as evolving agents to learn Hebbian-like learning rules for simple navigation tasks. However, until recently, neural plasticity has not been investigated with deep neural networks. 

Most recently, Miconi et. al \cite{miconi2018differentiable} demonstrated how neural plasticity can be tuned in deep neural networks together with other parameters using gradient decent. Inspired by the success of their system when extracting salient information cues, we propose the development of neural memory plasticity where the memory access mechanisms in neural memory models are made plastic to provide varying level of attention to the stored information, introducing a novel knowledge retrieval paradigm in NMNs. 

In a separate line of work, Harris et. al \cite{harris2019biologically} investigated the concept of a plasticity based working memory for visual recognition tasks. However, this is different from the proposed work as they are not utilising an external memory. In contrast, \cite{harris2019biologically} uses the neural plasticity itself as the temporal modelling mechanism. 

\subsection{Anomaly Detection}
In the domain of machine learning anomaly detection is primarily regarded as an unsupervised learning task. For instance, in \cite{eskin2002geometric} the authors try to detect anomalies in network traffic through a geometric pattern based framework. For video based anomaly detection, numerous methods including pixel-level features \cite{ermis2008motion}, trajectory features \cite{wu2010chaotic} and spatio-temporal features \cite{zhang2009detecting} are proposed which are subsequently categorised through an unsupervised learning paradigm. Numerous works have also exploited deep learning in an unsupervised setting for anomaly detection \cite{hasan2016learning, masci2011stacked,chong2017abnormal}. Please refer to \cite{kiran2018overview} for a complete review of these methods.  

In the medical domain, supervised anomaly detection methods have been preferred due to the inherent difficulties present in medical data. For instance, detecting abnormalities in an EEG recording is challenging in an unsupervised setting as abnormal artefacts are not clearly evident as there exist numerous natural variations among subjects. Therefore, what is defined as normal for one subject can be abnormal for another, requiring learning of abnormal and normal scenarios in order to discriminate. Hence most approaches in the medical domain have used supervised learning \cite{afshar2018brain,abiwinanda2019brain,roy2018deep,gumaei2019hybrid}.

With the recent spectacular success of deep learning methods for automatically learning task specific features, hand engineered features have been replaced by deep learned features for medical anomaly detection. Convolutional Neural Networks (CNNs) and recurrent neural networks such as LSTMs have been extensively applied to detect abnormal behaviour. However, as noted by \cite{yildirim2018deep,afshar2018brain} abnormalities can take many forms and there exists subtle differences between subjects, differences in the regions where the data is captured, etc. Hence we propose to augment the capacity of the modelling framework through the introduction of an external memory which can be used to store the observed knowledge and map the long-term dependencies between data samples. 

Recently, a neural memory network based approach for anomaly detection is proposed in \cite{gong2019memorizing} where the authors try to memorise the patterns within the normal data in order to detect abnormal instances. However, this approach is quite distinct from the proposed approach as we are learning our memory model from both normal and abnormal data and as such the memory learns to store distinctive characteristics from both normal and abnormal data streams. 

\section{Architecture}
In this section we introduce the structure of a typical NMN and its basic operations, and how they can be augmented to facilitate plasticity. 

\subsection{Neural Memory Networks}
\begin{figure}
\centering
\includegraphics[width=0.75\linewidth]{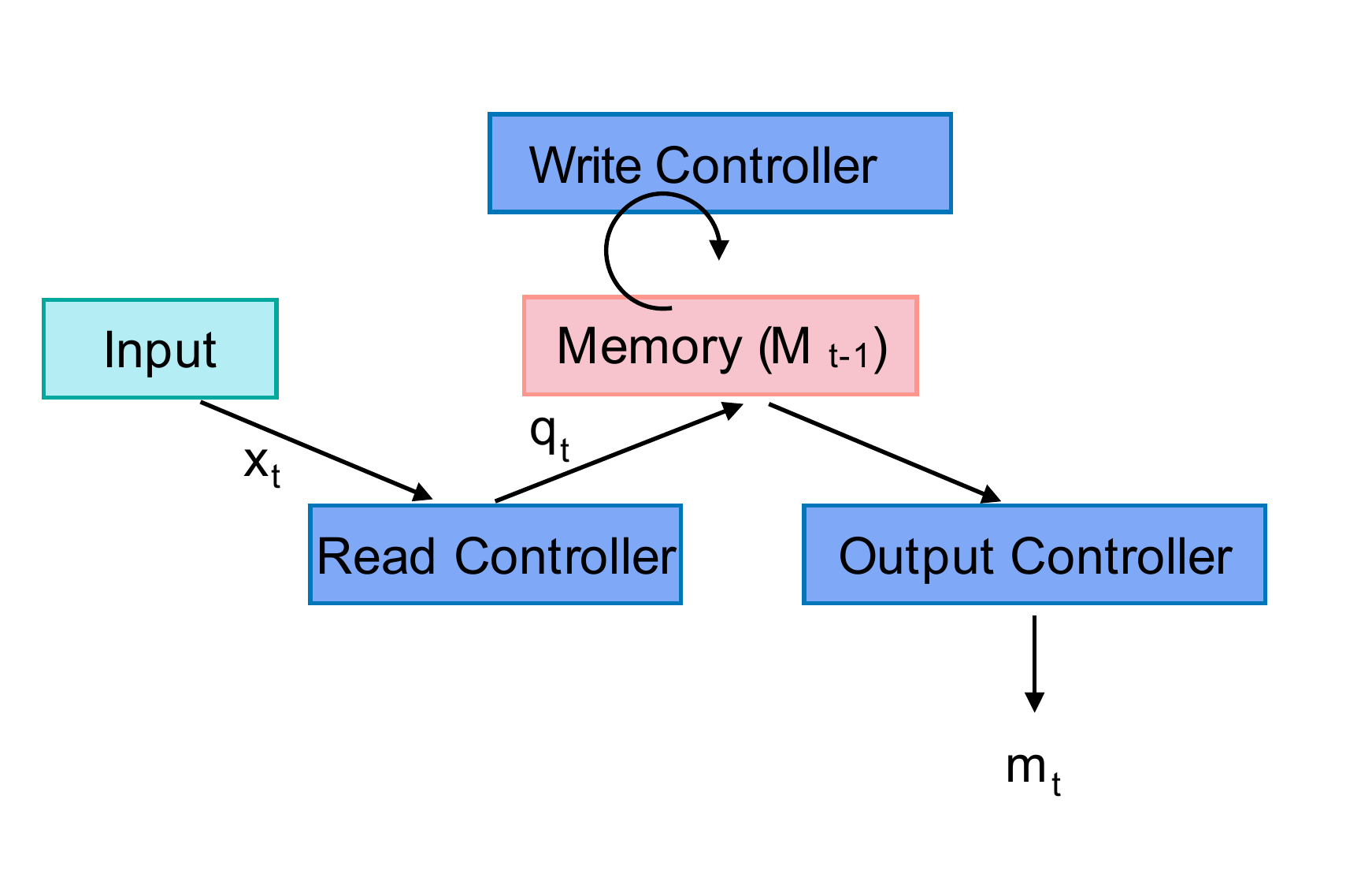}
\caption{Overview of an external memory which is composed of input, output and write controllers. The input controller determines what facts within the input are used to query the memory. The output controller determines what portion of the memory is passed as the output for that query. Finally the write controller updates the memory state and propagates it to the next time step.}
\label{fig:memory}
\end{figure}

As shown in Fig. \ref{fig:memory}, a typical memory module is composed of 1) a memory stack for information storage, 2) a read controller to query the knowledge stored in the memory, 3) a write controller for memory update, and 4) an output controller which controls what results are passed out from the memory. 

Let $M \in \mathbb{R}^{l \times k}$ be a memory stack with $l$ memory slots where each slot contains an embedding of dimension $k$. We represent the state of memory at time instance $t-1$ as $M_{t-1}$. In a typical memory implementation \cite{munkhdalai2017neural, fernando2018task, fernando2018learning} first the read controller passes the input, $x_t$, at the current time instance $t$ through a read function composed of a Long Short Term Memory (LSTM) cell such that, 
\begin{equation}
q_t = f^{LSTM}_r (x_t),
\end{equation}

Following \cite{bahdanau2014neural}, using a softmax function we quantify the similarity between the content stored in each slot of $M_{t-1}$ and the query vector, $q_t$, such that,
\begin{equation}
z_t = \textrm{softmax}({q_t}^\top M_{t-1}).
\end{equation}

Now the output controller can retrieve the memory output, $m_t$, for the current state by,
\begin{equation}
 m_t=z_{t}^\top M_{t-1}.
\label{eq:memory_out}
\end{equation}

%
%

%

Finally the write controller, which also uses an LSTM cell generates an update vector, $m'_t$, to update the memory,
\begin{equation}
m'_t = f_w^{LSTM} (m_t),
\end{equation}
and updates the memory using,
\begin{equation}
M_t = M_{t-1} (I - z_{t} \otimes e_k)^\top + (m'_t \otimes e_l) (z_{t} \otimes e_k)^\top,
\end{equation}
where $I$ is a matrix of ones, $e_l \in \mathbb{R}^l$ and $e_k \in \mathbb{R}^k$ are vectors of ones and $\otimes$ deontes the outer product which duplicates its left vector $l$ or $k$ times to form a matrix. 

\subsection{Injection of Plasticity for Memory Components}
We follow the formulation of the Hebbian rule proposed in \cite{miconi2018differentiable} for its flexibility and simplicity. 

We define a fixed component and plastic component for each pair of neurons $i$ and $j$, and the plastic component is stored in a Hebbian trace $\mathrm{Hebb_{i,j}}$, which evolves over time based on the inputs and outputs. 

Formally let there be two input layers, each with $k$ neurons and let $w \in \mathbb{R}^{k \times k}$ be the fixed weights and $\mathrm{Hebb} \in \mathbb{R}^{k \times k}$ store the Hebbian trace. Then a sample input $x_t$ to the first layer at time instance, $t$, is passed to the next layer such that, 

\begin{equation}
x^j_{t+1} = \mathrm{tanh}( \sum_{\forall_{i,j} \in k }w^{i,j} x^i_t + \alpha^{i,j}\hspace{1px}\mathrm{Hebb}^{i,j}_{t+1}\hspace{1px}x^i_t  ),
\label{eq:plasticity}
\end{equation}
where $x^i_t$ denotes the $i^{th}$ embedding in input $x_t$ and,
\begin{equation}
\mathrm{Hebb}^{i,j}_{t+1} = \mathrm{Hebb}^{i,j}_t + \eta \hspace{1px} x^j_t (x^i_{t-1} - x^j_t \hspace{1px} \mathrm{Hebb}_t^{i,j}),
\end{equation}

where $\alpha$ is a coefficient which controls the contribution from fixed and plastic terms of a particular weight connection and $\eta$ is the learning rate of plastic components.

Thus using the formulation in Eq. \ref{eq:plasticity} we replace the components of the read, write and output controllers to facilitate plasticity such that,
\begin{equation}
q_{t} = \mathrm{tanh}( \sum_{\forall_{i,j} \in k } \dot{w}^{i,j} x^i_{t-1} + \dot{\alpha}^{i,j}\hspace{1px}\mathrm{\dot{Hebb}}^{i,j}_{t}\hspace{1px}x^i_{t-1}  ),
\label{eq:plasticity_2}
\end{equation}

\begin{equation}
z_t = \textrm{softmax}({q_t}^\top M_{t-1}),
\end{equation}

\begin{equation}
\beta_t = {z_t}^\top M_{t-1},
\end{equation}


\begin{equation}
m_{t} = \mathrm{tanh}( \sum_{\forall_{i,j} \in k } \hat{w}^{i,j} \beta^i_{t-1} + \hat{\alpha}^{i,j}\hspace{1px}\mathrm{\hat{Hebb}}^{i,j}_{t}\hspace{1px}\beta^i_{t-1}  ),
\label{eq:plasticity_3}
\end{equation}

\begin{equation}
m'_{t} = \mathrm{tanh}( \sum_{\forall_{i,j} \in k } \tilde{w}^{i,j} m^i_{t-1} + \tilde{\alpha}^{i,j}\hspace{1px}\mathrm{\tilde{Hebb}}^{i,j}_{t}\hspace{1px}m^i_{t-1}  ).
\label{eq:plasticity_3}
\end{equation}

\subsection{Abnormality Detection}
\label{sec:abnormality_detection}
\begin{figure*}[htbp]
\centering
\includegraphics[width=.8\textwidth]{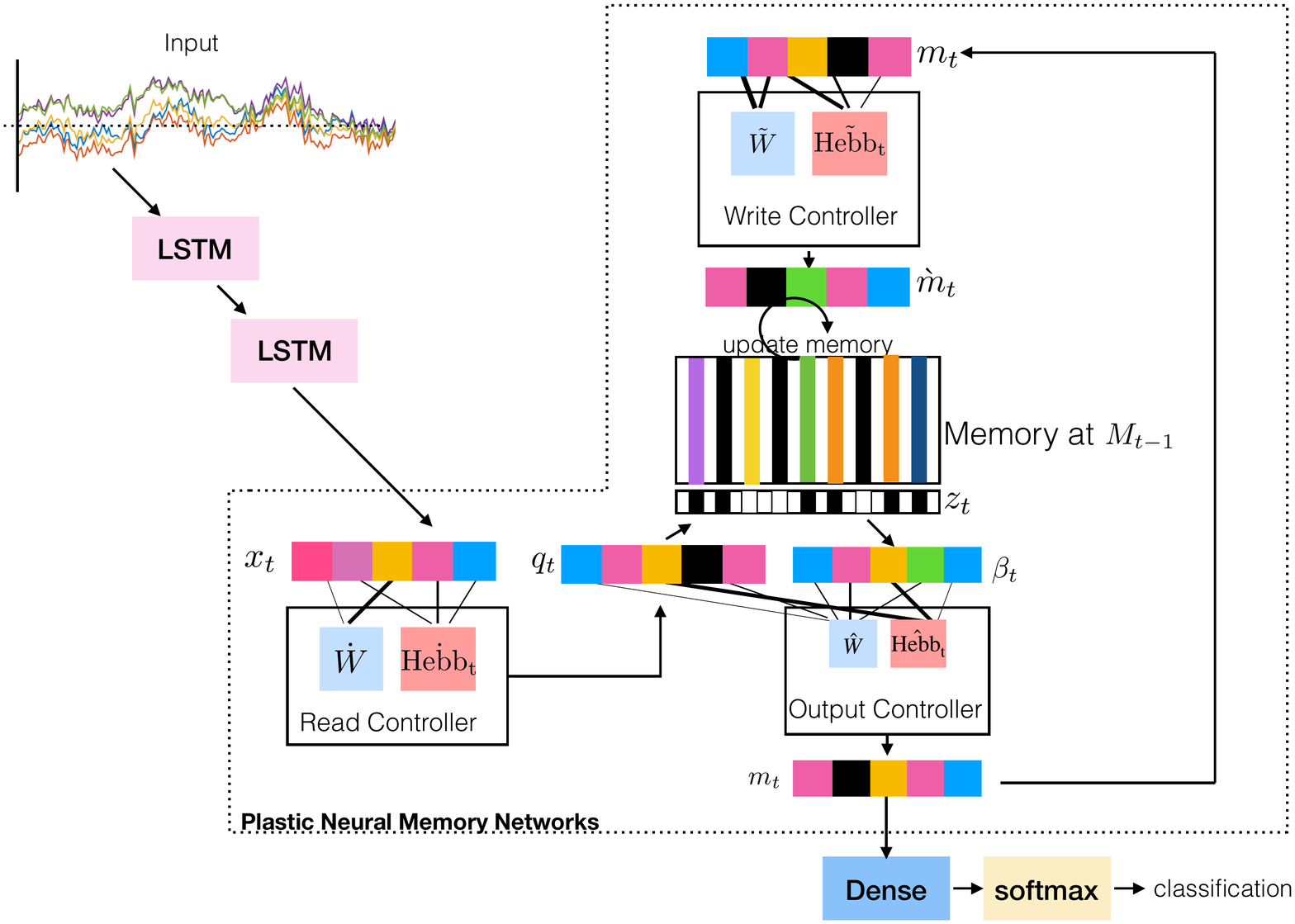}
\caption{Overview of the Proposed Abnormality Detection Framework with proposed Plastic NMN: We map the input sequences using two layers of LSTMs. The resultant embedding is utilised to retrieve the salient information from the stored knowledge in the memory and the retrieved vector is passed through the output and write controllers which determines the memory output at the current time instance and how to update the memory, respectively. These controllers utilise a combination of fixed weights and plastic components. A dense layer with softmax activation is used to determine the classification of the input. }
\label{fig:full_model}
\end{figure*}

To evaluate the abnormality detection accuracy of the proposed memory architecture we conduct three experiments. Two experiments are conducted with EEG data and one with MRI data. Fig. \ref{fig:full_model} illustrates the structure of the proposed model. 

\begin{figure}[htbp]
\includegraphics[width=\linewidth]{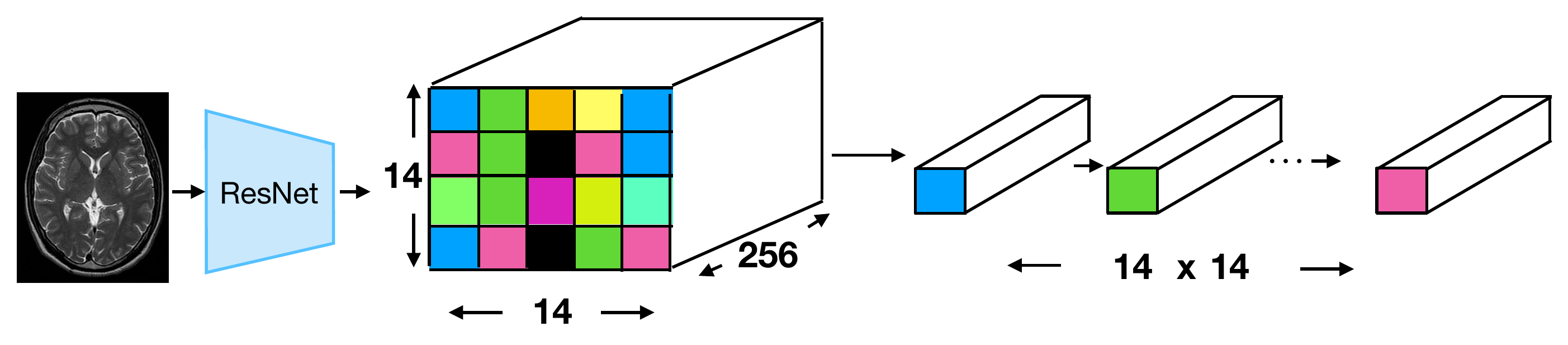}
\caption{Feature Extraction from MRI data: We utilise a ResNet \cite{he2016deep} CNN architecture pre-trained on ImageNet \cite{deng2009imagenet} as our visual feature extractor, and extract features from the ``Activation-85'' layer which has an output dimensionality of $14 \times 14 \times 256$. Then we consider each element in the extracted feature block individually and map that to a sequence, row wise, from top left to bottom right, which containes $14 \times 14 = 196$ elements. }
\label{fig:image_features}
\end{figure}

For the EEG data we consider a short time window of the EEG signals, hence requiring temporal modelling within the time window considered. In the MRI experiment, as the MRI is a spatial representation, motivated by \cite{xiong2016dynamic} we extract spatial features from the MRI using a pre-trained ResNet 50 \cite{he2016deep} and represent each element of the extracted feature block as an element in a sequence. The process is illustrated in Fig. \ref{fig:image_features}. 

This allows us to analyse the correspondences between pixels. For modelling the short-term relationships within the sequence we use LSTMs. For extracting out the relevant attributes through long-term dependencies we employ the proposed memory architecture. The normal/ abnormal classification or the tumour type classification (i.e 3 classes, Meningioma, Pituitary, and Glioma) is generated through a dense layer with softmax classification.  

As opposed to unsupervised abnormality detection, which is frequently used in video tasks, we follow the baseline algorithms in the medical domain and use supervised learning, enabling direct comparison. 

\section{Evaluations}
The following subsections discuss the dataset details, experimental setup and results of the experiments conducted for abnormal EEG detection, MRI tumour type classifications and EEG based schizophrenia risk detection. 

\subsection{Abnormal EEG Identification}
\label{sec:abnormal_EEG_Detection}
\subsubsection{Dataset} For this experiment we use the TUH Abnormal EEG database \cite{lopez2015automated} which is the world's largest publicly available dataset of its type \cite{roy2018deep}. This dataset consists of 1488 abnormal EEGs and 1529 normal EEGs and is demographically balanced with respect to the age and gender of patients. For training and testing of the systems we utilise the splits provided by the authors of the dataset. 

The EEG signals were obtained at 250 HZ and we extract 60 second samples (i.e 1500 data points within a window) using a sliding window approach with 50\% overlap between two consecutive windows. Similar to \cite{yildirim2018deep} we utilise only the T5-O1 channel of the EEG recordings as input to our model. 

We perform min-max scaling of the input and no other pre-processing is performed. The final decision for the entire EEG is obtained through majority voting.  

\subsubsection{Experimental Setup}\label{sec:experimental_setup} From the training set we use an 80\% - 20\% split for training and validation. We train the model using the Adam \cite{kingma2014adam} optimiser and binary cross entropy loss for 50 epochs. Similar to \cite{miconi2018differentiable} we share the same values for all $\eta$s (i.e $\dot{\eta}, \hat{\eta} $ and $\tilde{\eta}$). For both LSTMs we maintain the same embedding dimension $k$ which is also used as the embedding dimension of the memory.

\begin{figure*}[htbp]
\centering
 \subfloat[][$l$ vs Accuracy]{\includegraphics[width=.3\textwidth]{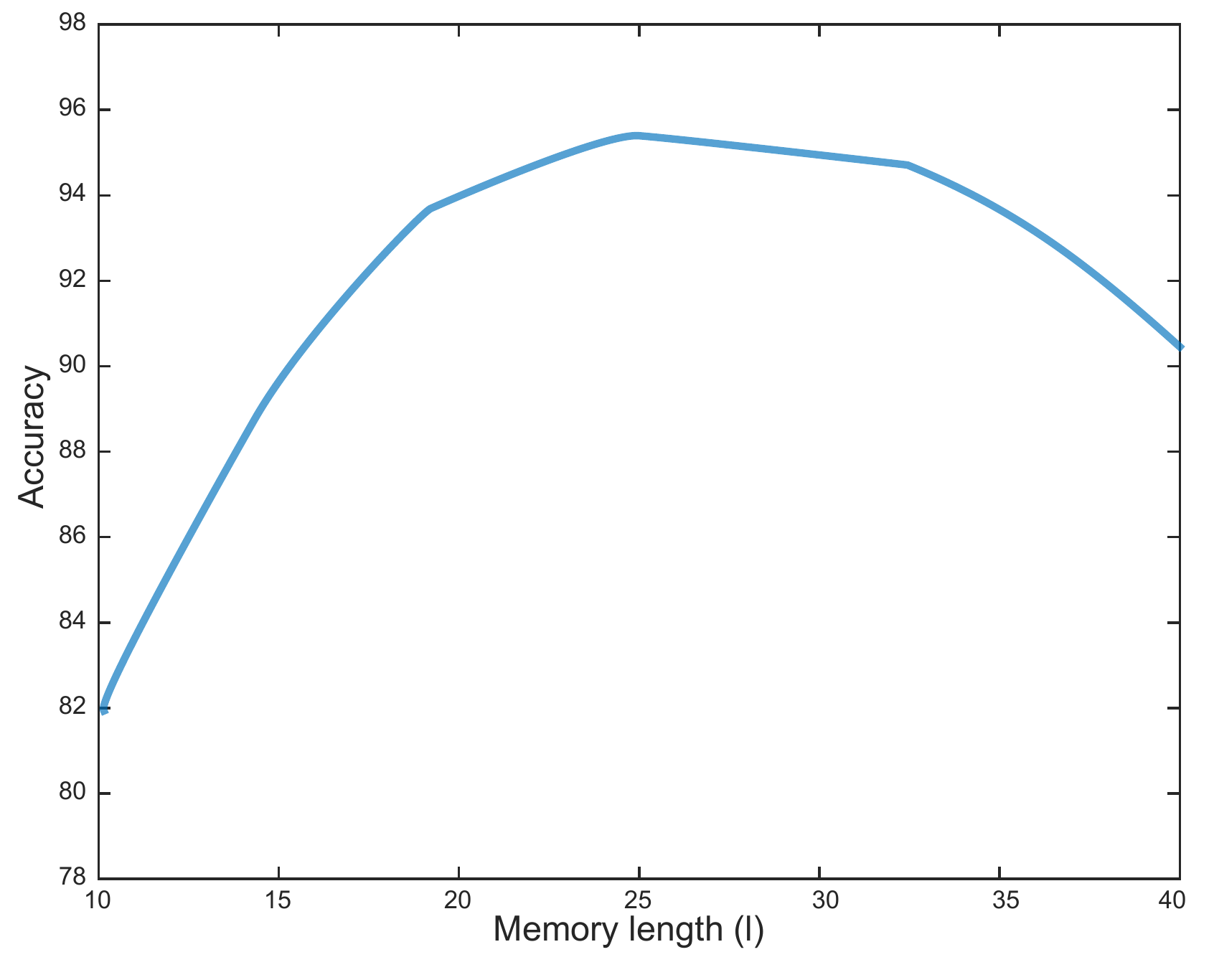}}
 \subfloat[][$k$ vs Accuracy]{\includegraphics[width=.3\textwidth]{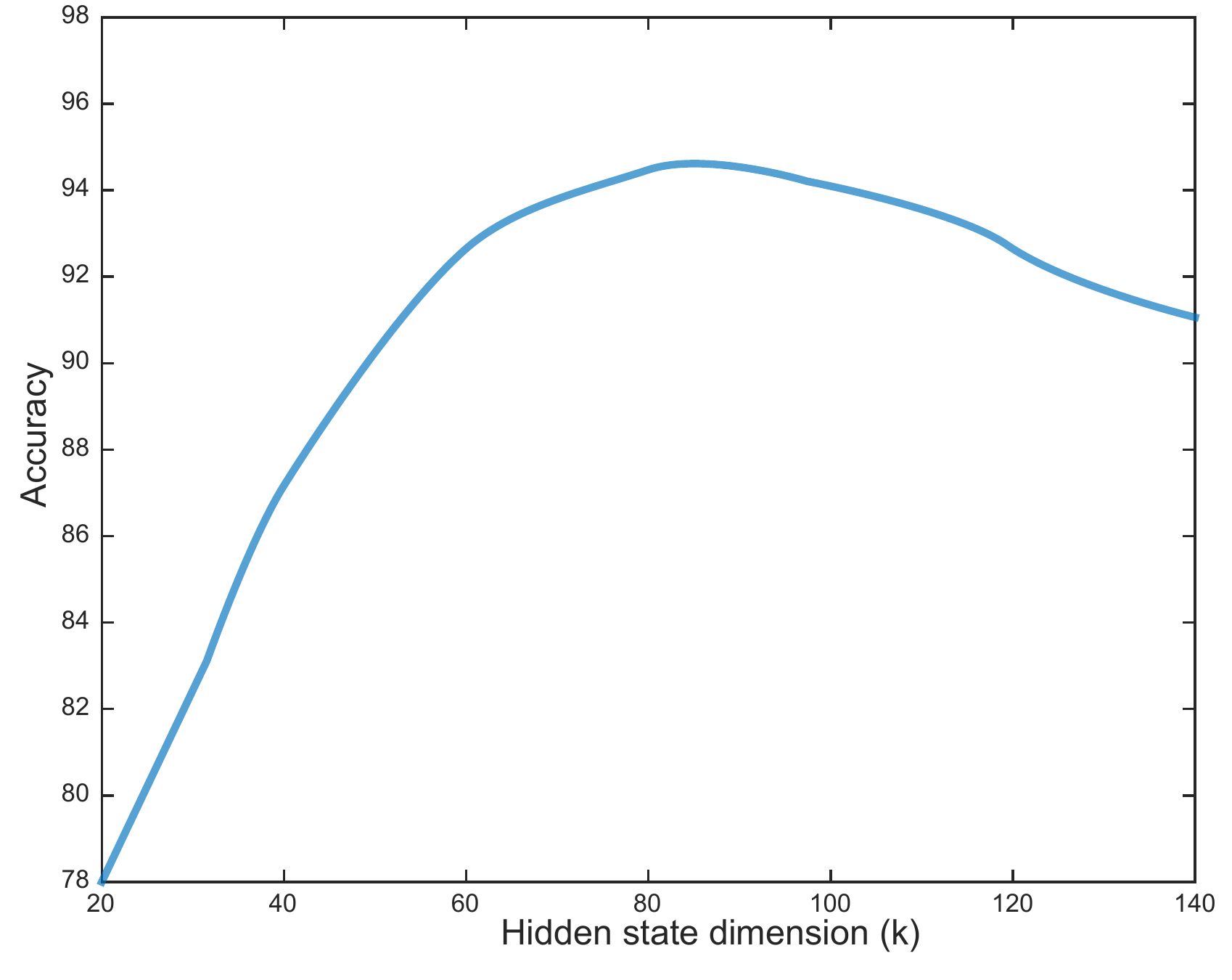}}
 \subfloat[][$\eta$ vs Accuracy]{\includegraphics[width=.295\textwidth]{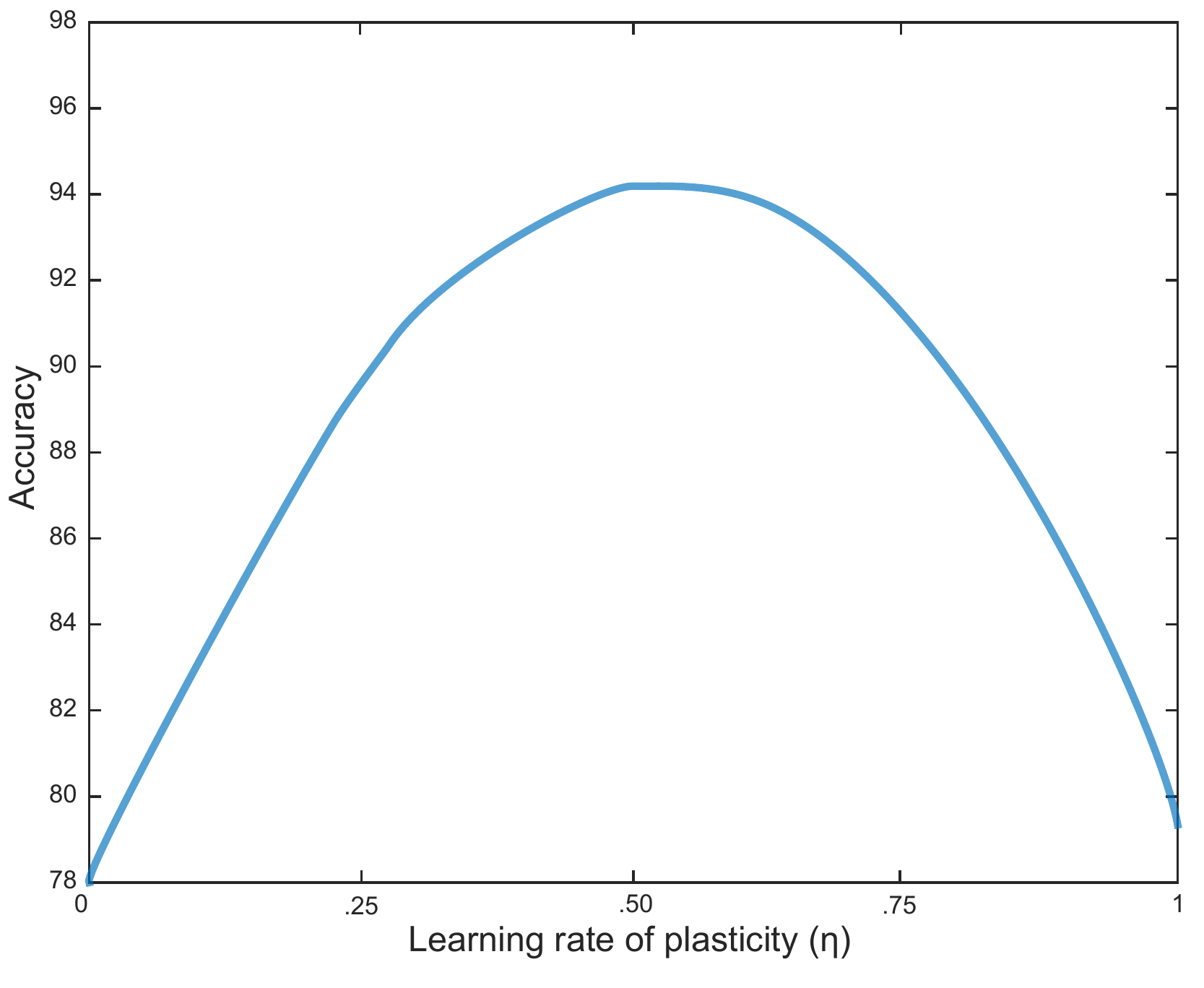}}
  \caption{Hyper-parameter evaluation for the Abnormal EEG Evaluation (see Sec. \ref{sec:abnormal_EEG_Detection}). System performance as a single parameter ($l$, $k$ or $\eta$) is changed while the others are held constant is reported. Values we selected for the three parameters in this experiment are $k=80$, $l=25$, and $\eta=0.50$}
\label{fig:ex_1_hype}
\end{figure*}

Hyper-parameters $k$, $l$, and $\eta$ are evaluated experimentally using the validation set and these evaluations are shown Fig. \ref{fig:ex_1_hype}. As $k=80$, $l=25$, and $\eta=0.50$ provides best accuracies in the validation set we use these parameters for model training. 

\subsubsection{Results}\label{sec:abnormal_eeg_results} Experimental results are presented in Tab. \ref{tab:tab_1}. For comparisons we report the results of the k-Nearest Neighbour (kNN) and Random Forest (RF) classifiers utilised by \cite{lopez2015automated}, where the authors first apply class dependent PCA to extract features from the EEG window and train the kNN and RF classifiers using those features. In \cite{Lopez_ms_thesis} the authors apply a 2D CNN model on four channels of the EEG signal. In \cite{yildirim2018deep}, the authors propose a 1D CNN model that uses the T5-O1 channel of the EEG as the input. This model is extended in \cite{roy2018deep} where the authors combine of one-dimensional convolution layers together with GRU layers. To further illustrate the utility of the proposed plasticity based memory addressing mechanism we evaluate the performance of the baseline memory model proposed in \cite{munkhdalai2017neural}. For this model, similar to the proposed abnormality detection model in Sec. \ref{sec:abnormality_detection}, we use 2 layers of LSTMs followed by the memory model in \cite{munkhdalai2017neural} and a dense layer with softmax activation for generating the final classification. The hyper-parameters of this model, memory length, $l$, and embedding dimension, $k$, are evaluated experimentally using the validation set and we set $l=25$ and $k=50$. 

\begin{table}[htbp]
\centering
\caption{Evaluations of abnormal EEG identification using the dataset proposed in \cite{lopez2015automated}}
\resizebox{.5\linewidth}{!}{
\begin{tabular}{|c|c|}
\hline
Method   & Accuracy      \\ \hline
kNN \cite{lopez2015automated}     & 58.2          \\ \hline
RF \cite{lopez2015automated}      & 68.3          \\ \hline
2D CNN \cite{Lopez_ms_thesis}   & 78.8          \\ \hline
1D CNN \cite{yildirim2018deep}  & 79.4          \\ \hline
1D-CNN-RNN \cite{roy2018deep} & 82.27 \\ \hline
Memory model of  \cite{munkhdalai2017neural}  & 83.4          \\ \hline  \hline
Plastic NMN (Proposed) & \textbf{93.2} \\ \hline
\end{tabular}}
\label{tab:tab_1}
\end{table}

When comparing the results we observe a significant performance boost with the introduction of deep learning techniques compared to traditional classifiers such as kNN and RF. With the introduction of GRU layers the performance of the 1D-CNN model is improved in \cite{roy2018deep}. Though we observe a slight increase in performance with the addition of a memory component in the memory model of \cite{munkhdalai2017neural}, we do not observe a substantial accuracy increase mainly due to the limitations of the memory read and update mechanisms. However, with the proposed memory model we provide the utility of long-term dependency modelling and allow the model to extract the most salient components for decision making using the proposed plastic components in the memory addressing mechanism. This allows us to outperform all the baseline models by a significant margin. 

\subsection{MRI Tumour Type Classification}
\label{sec:mri}
\subsubsection{Dataset} We use the MRI database provided by \cite{cheng2015enhanced}. The dataset is comprised of 3064 brain tumour MRI images taken from 233 patients. Types of brain tumours in the dataset are meningioma (1426 samples), glioma (708 samples), and pituitary (930 samples).

\subsubsection{Experimental Setup}\label{sec:mri_ex_setup} 
\begin{figure*}[htbp]
\centering
 \subfloat[][$l$ vs Accuracy]{\includegraphics[width=.3\textwidth]{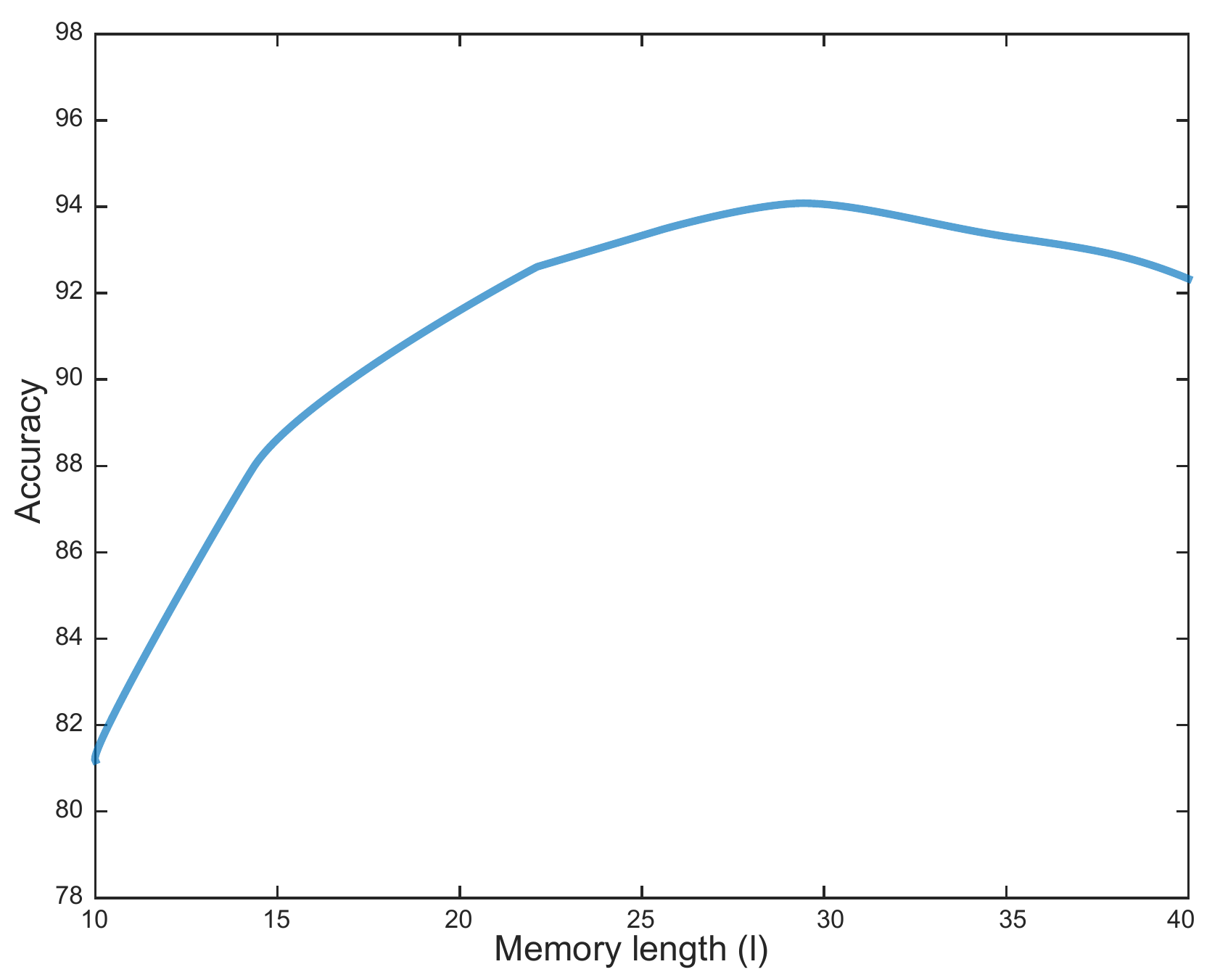}}
 \subfloat[][$k$ vs Accuracy]{\includegraphics[width=.3\textwidth]{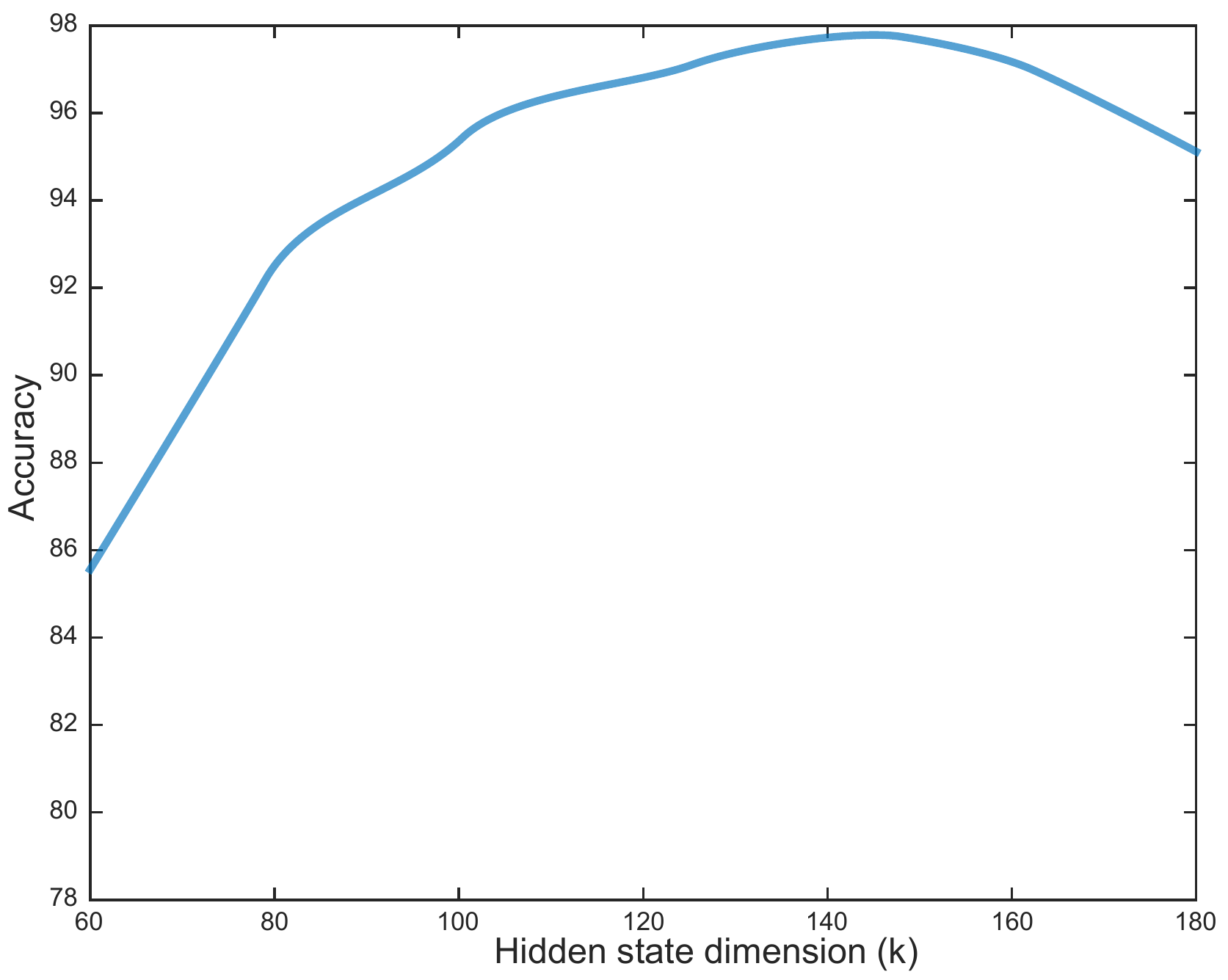}}
  \subfloat[][$\eta$  vs Accuracy]{\includegraphics[width=.3\textwidth]{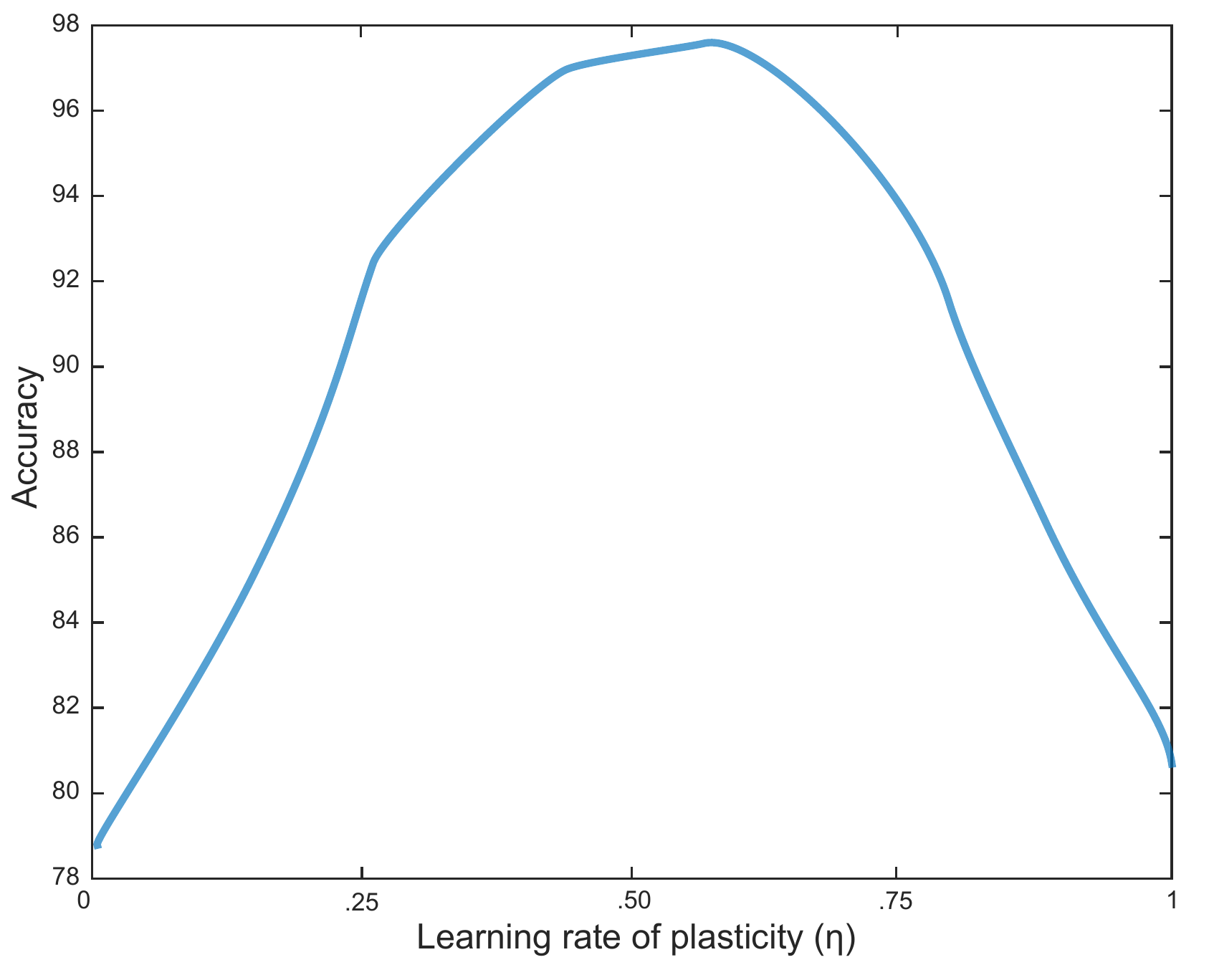}}
  \caption{Hyper-parameter evaluation for the MRI Tumour Type Classification Evaluation (see Sec. \ref{sec:mri}). System performance as a single parameter ($l$, $k$ or $\eta$) is changed while the others are held constant is reported. Values we selected for the three parameters in this experiment are $k=150$, $l=30$, and $\eta=0.55$}
\label{fig:ex_2_hype}
\end{figure*}

Due to the small dataset size we perform data augmentation as per \cite{sajjad2019multi}, where we apply random rotations and flipping to augment the data. Then using a ResNet 50 model \cite{he2016deep} pre-trained on ImageNet \cite{deng2009imagenet} we extract features from the Activation 85 layer. Then, as shown in Fig. \ref{fig:image_features} we generate a sequence of features which are used as the input to the proposed model. As there is no standard training/ testing splits for this dataset, similar to \cite{gumaei2019hybrid}, we apply 5-fold cross validation where 80\% of the training data is used for model training while the remaining 20\% is used for validation. We train the model using the Adam \cite{kingma2014adam} optimiser with categorical cross entropy loss for 50 epochs.  Similar to Sec. \ref{sec:experimental_setup} we evaluate hyper-parameters $k$, $l$, and $\eta$ experimentally and show these evaluations in Fig. \ref{fig:ex_2_hype}. Based on this evaluation we set $k=150$, $l=30$, and $\eta=0.55$.

\subsubsection{Results} We present the evaluation results in Tab. \ref{tab:tab_2} where we compare the proposed method against a series of baselines. In \cite{sajjad2019multi} the authors utilise a pre-trained VGG-19 architecture and fine-tune it for the classification task. In \cite{abiwinanda2019brain} the authors use a shallow CNN architecture which comprises a single convolution layer. In \cite{afshar2018brain} the authors investigate the utility of capsule networks for exploiting the spatial relationships within the image. In \cite{gumaei2019hybrid} the authors propose the use of the GIST descriptor with PCA for feature extraction as opposed to using deep learned features. In addition, we utilise the baseline memory model of  \cite{munkhdalai2017neural} in our comparisons. The structure of this model is as defined in Sec. \ref{sec:abnormal_eeg_results} and we have evaluated the hyper-parameters $l$ and $k$ experimentally and values of $l=30$ and $k=120$ are chosen.

\begin{table}[htbp]
\centering
\caption{Evaluations of MRI tumour type classification using the dataset proposed in \cite{cheng2015enhanced}}
\resizebox{.5\linewidth}{!}{
\begin{tabular}{|c|c|}
\hline
Method   & Accuracy      \\ \hline
CNN \cite{abiwinanda2019brain}      & 84.19          \\ \hline
CapsuleNet  \cite{afshar2018brain}     & 86.56          \\ \hline
GIST PCA  \cite{gumaei2019hybrid} & 92.61          \\ \hline
CNN  \cite{sajjad2019multi} & 94.46          \\ \hline
Memory model of  \cite{munkhdalai2017neural}  & 90.1    \\ \hline  \hline 
Plastic NMN (Proposed) & \textbf{97.54} \\ \hline
\end{tabular}}
\label{tab:tab_2}
\end{table}

When comparing the results it is clear that there exists a higher sensitivity within the model performance, especially among the deep learned models, due to data scarcity. Even though we expect the baseline memory model of \cite{munkhdalai2017neural} to obtain better performance compared to other deep learning baselines, it has obtained poorer performance, mainly due to the higher dimensionality of the MRI data, where a better memory access scheme is required to retrieve the most salient information. In contrast, the proposed method, exploiting the neural plasticity of the knowledge retrieval process, has been able to optimally utilise the limited training data that is available and effectively capture the salient features for the task at hand. 

\subsection{EEG based Schizophrenia Risk Detection} 
\label{sec:eeg_schizophrenia}
\subsubsection{Dataset} We use the EEG recordings from the auditory oddball trials conducted in \cite{bruggemann2013mismatch} where the subjects listen to different tones of which some are frequent and some are less frequent. Several studies in neuroscience research have indicated a reduction in the amplitude of brain response to auditory change detection in patients who risk development of schizophrenia \cite{moghaddam2012revolution}, and a number of studies have subsequently employed the auditory oddball paradigm for detection of schizophrenia \cite{shin2009pre, shin2011aberrant, bodatsch2011prediction}. 

The dataset includes EEG recordings from children aged 9 to 12 years, including 65 children with an increased Risk of Schizophrenia (RSz) due to a positive family history of schizophrenia (in at least one first- or second-degree relative) and/or their presenting multiple replicated developmental antecedents of schizophrenia, and 39 Typically Developing (TD) children who presented none on those developmental antecedents or family history of schizophrenia in first-, second-, or third-degree relatives. Stimuli used for the auditory oddball paradigm were 1600 tones at 1000 Hz, including 1360 (85\%) standard tones of 25ms duration and 240 (15\%) deviant tones of 50ms duration. These standard and deviant tones were presented in pseudo-random order to avoid successive deviant stimuli, with an isochronous inter-stimulus interval of 300ms. The participants passively listen to the auditory oddball task and their electrocortical data are recorded according to the international standard 10-10 system of electrode placement. Please refer to \cite{laurens2010error, bruggemann2013mismatch} for details.

We select the Fz, FCz, Cz, CPz and Pz channels from the EEG recordings. As pre-processing we apply a 50 Hz notch filter and artefact rejection to the ocular channels to remove blinks \cite{bruggemann2013mismatch}. After the occurrence of the stimulus (i.e standards/ deviants) we extract a 300ms window from each of the selected channels of the EEG and perform min-max scaling of each channel separately. No other pre-processing is performed. 

\subsubsection{Experimental Setup}\label{sec:asz_ex_setup} Due to the unavailability of standard training and testing splits, we adopt a 5 fold cross validation of the training set where we select 80\% of the training data for training the model and 20\% for validation. In order to generate the final classification of each participant (i.e RSz / TD) we utilise majority voting. 

\begin{figure*}[htbp]
\centering
 \subfloat[][$l$ vs Accuracy]{\includegraphics[width=.3\textwidth]{figures/hyper-params/ex_3_l.pdf}}
 \subfloat[][$k$ vs Accuracy]{\includegraphics[width=.3\textwidth]{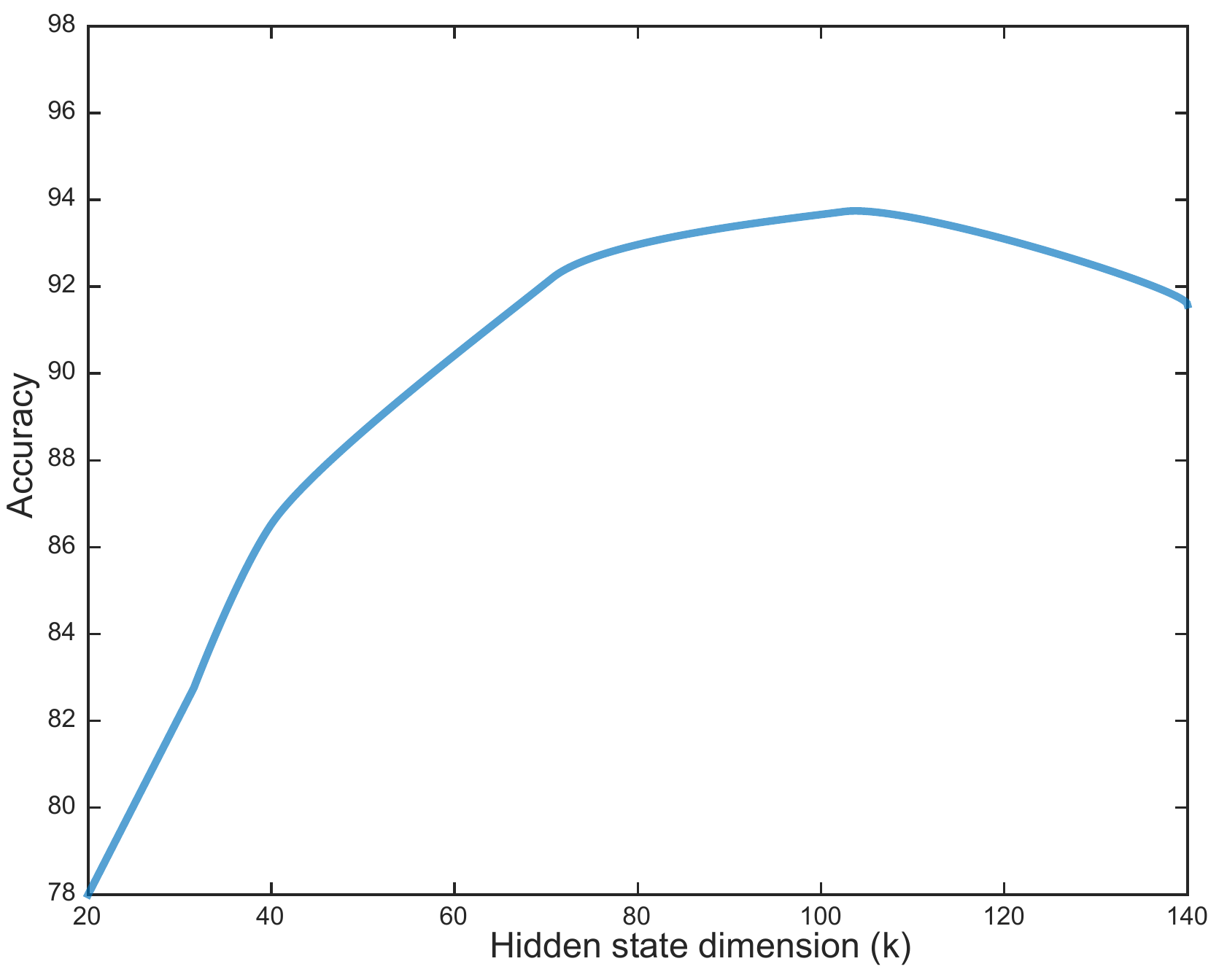}}
  \subfloat[][$\eta$  vs Accuracy]{\includegraphics[width=.295\textwidth]{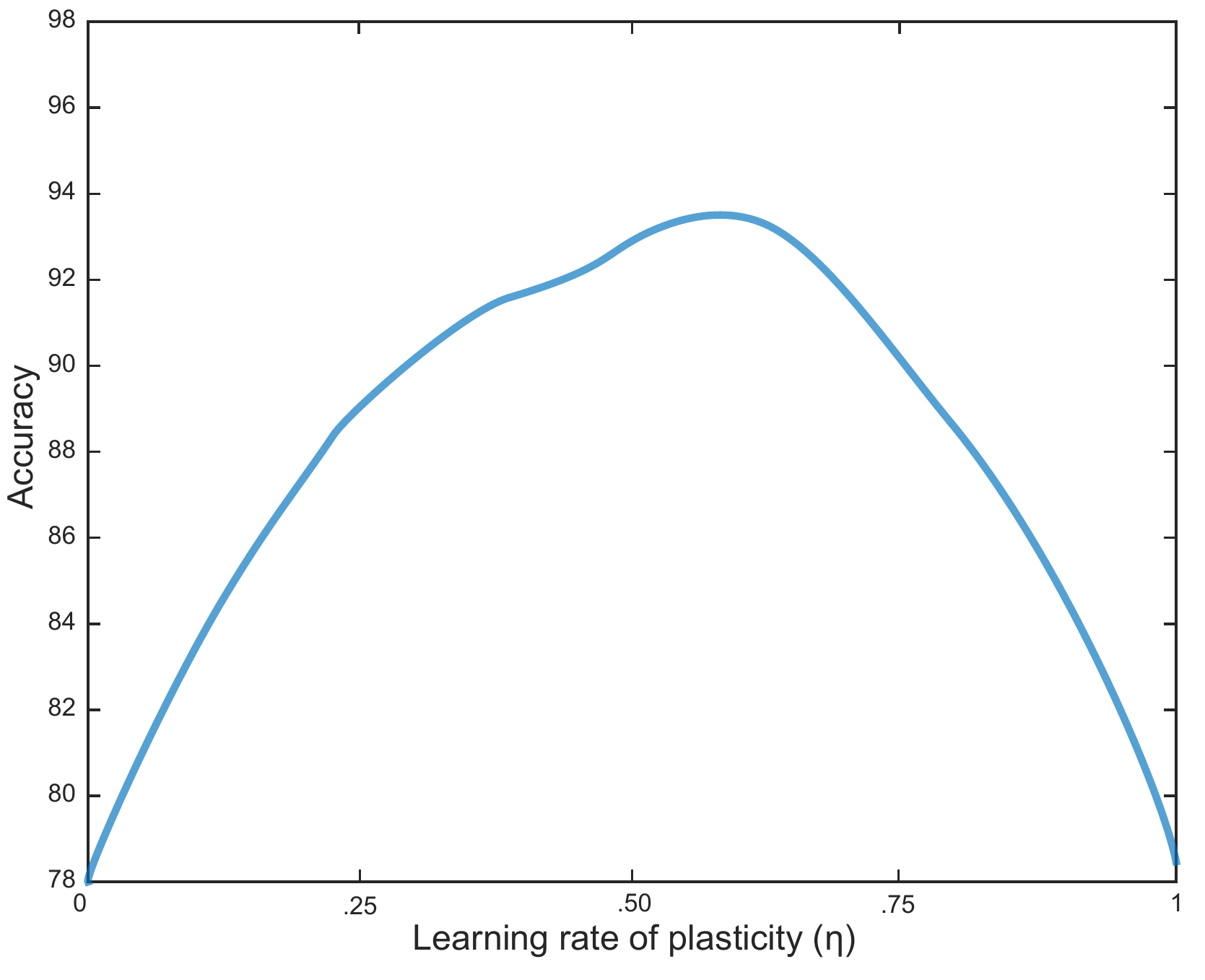}}
  \caption{Hyper-parameter evaluation for the EEG Based Schizophrenia Risk Detection Evaluation (see Sec. \ref{sec:eeg_schizophrenia}). System performance as a single parameter ($l$, $k$ or $\eta$) is changed while the others are held constant is reported. Values we selected for the three parameters in this experiment are $k=100$, $l=30$, and $\eta=0.65$}
\label{fig:ex_3_hype}
\end{figure*}

Similar to Sec. \ref{sec:experimental_setup} we evaluate hyper-parameters $k$, $l$,  and $\eta$ experimentally and present these evaluations in Fig. \ref{fig:ex_3_hype}. As $k=100$, $l=30$, and $\eta=0.65$ provides best accuracies we use these parameters for model training. 

\subsubsection{Results} Evaluation of the proposed method along with the baselines are reported in Tab. \ref{tab:tab_3}. To the best of our knowledge there are no existing machine learning models that attempt classification of schizophrenia risk using EEGs. 

The authors of \cite{bruggemann2013mismatch} evaluate the peak amplitudes elicited by the mismatch between the deviant and standard tones in the auditory oddball paradigm and demonstrate that there exists a statistical difference between children at risk for schizophrenia relative to typically developing children. Hence as the first baseline model we extract the early positive (between 160 - 290 ms after the stimulus) and early negative (between 80 - 200 ms after the stimulus) peak amplitude values which are baselined to the average amplitude of the 100 ms window before the stimulus; and these features are subsequently passed through a SVM classifier. 

In order to evaluate the performance of a deep learned model we pass the 5 input channels of the EEG through a single 2D convolution layer with 32 kernels, each with the kernel size of 64 $\times$ 5 \footnote{this kernel size is experimentally evaluated}. The resultant feature vector is passed through an LSTM and the final classification is obtained by a dense layer with softmax activation. Finally, we use the baseline memory model of  \cite{munkhdalai2017neural} where the structure of this model is defined in Sec. \ref{sec:abnormal_eeg_results} and hyper-parameters $l$ and $k$ are evaluated using the validation set and are set to $l=25$ and $k=100$.

\begin{table}[htbp]
\centering
\caption{Evaluations of EEG based schizophrenia detection task using the dataset proposed in \cite{moghaddam2012revolution}}
\resizebox{.5\linewidth}{!}{
\begin{tabular}{|c|c|}
\hline
Method   & Accuracy      \\ \hline
SVM      &    43.88      \\ \hline
CNN & 76. 33          \\ \hline  
Memory model of  \cite{munkhdalai2017neural}  & 78.5    \\ \hline 
Plastic NMN (Proposed) & \textbf{93.85} \\ \hline
\end{tabular}}
\label{tab:tab_3}
\end{table}

When comparing the results it is evident that use of the standard peak amplitude feature is not sufficient to obtain good segregation between the RSz and TD groups. Furthermore, baseline CNN and memory models have not been able to generate satisfactory performance. We believe this is due to the inherent challenges in the task as there is less clear separation between the RSz and TD groups. However, through the utilisation of the proposed memory model and via augmenting the read and write mechanisms, we are able to attain superior classification results.  

\section{Discussion}

In this section we provide qualitative evidence indicating the importance of neural plasticity in the memory addressing mechanism and interpret what the model has learnt in terms of model activations. 

\subsection{Importance of Neural Plasticity in NMNs}
\label{sec:discuss_1}
We extract the output of the memory model in \cite{munkhdalai2017neural} (i.e Eq. \ref{eq:memory_out}) and the output of the proposed memory model (i.e Eq. \ref{eq:plasticity_3}) for the experiment outlined in Sec. \ref{sec:eeg_schizophrenia}. The model in \cite{munkhdalai2017neural} utilises attention to extract relevant information from the stored knowledge in the neural memory while the proposed method exploits a combination of fixed weights and neural plasticity components. 

\begin{figure*}[htbp]
\centering
 \subfloat[][]{\includegraphics[width=.43\textwidth]{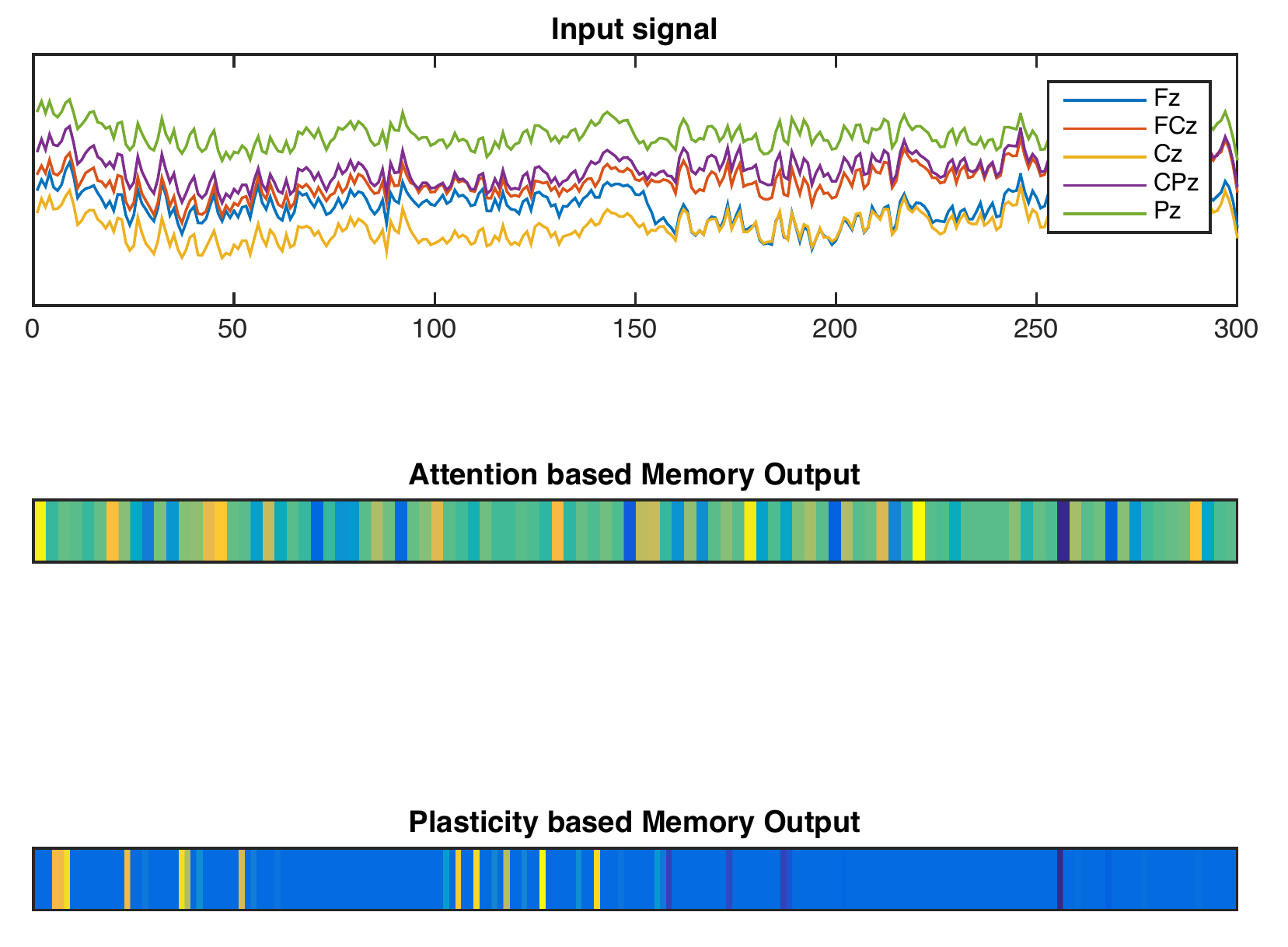}}
 \subfloat[][]{\includegraphics[width=.43\textwidth]{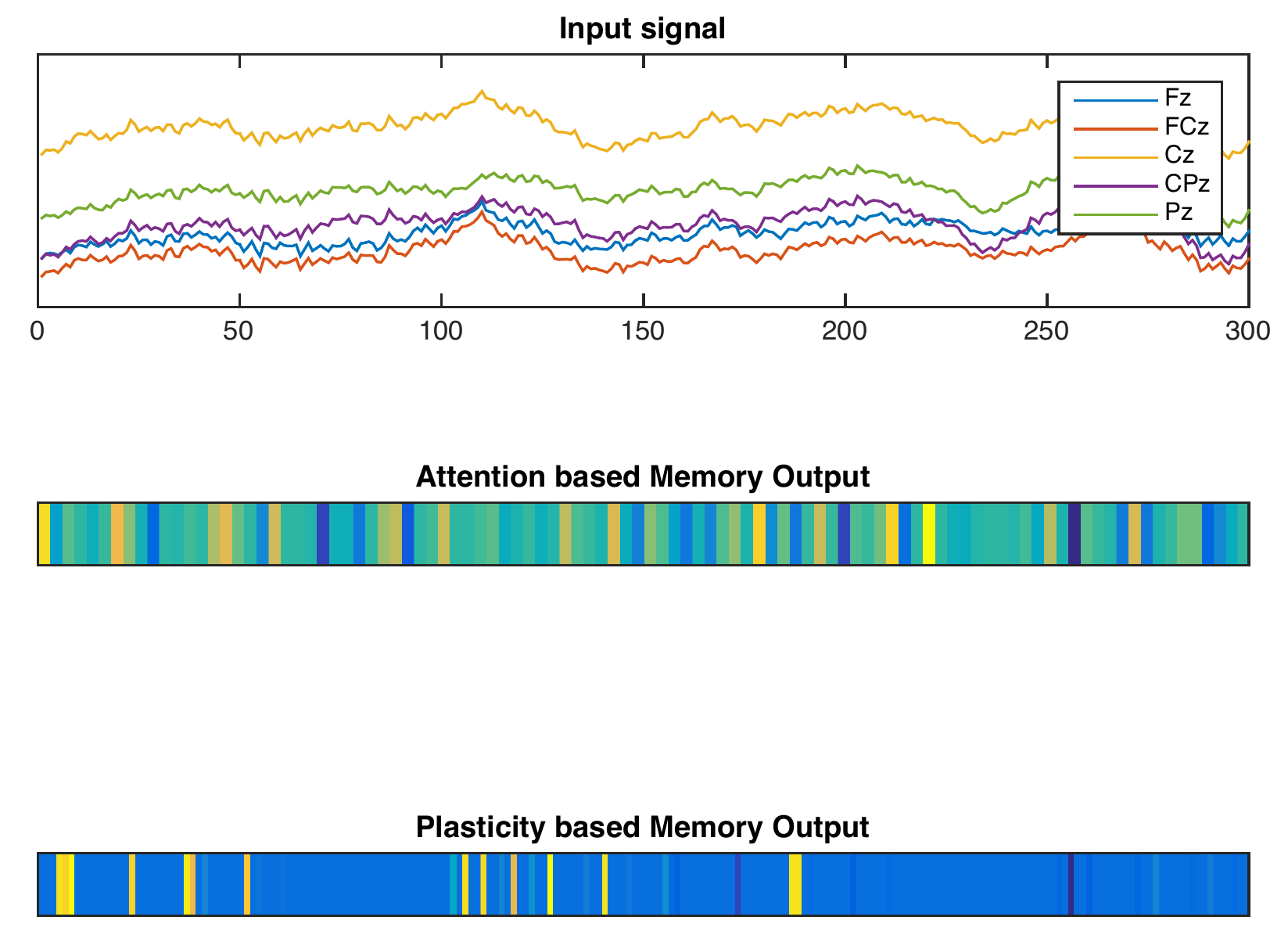}} \\
 \subfloat[][]{\includegraphics[width=.42\textwidth]{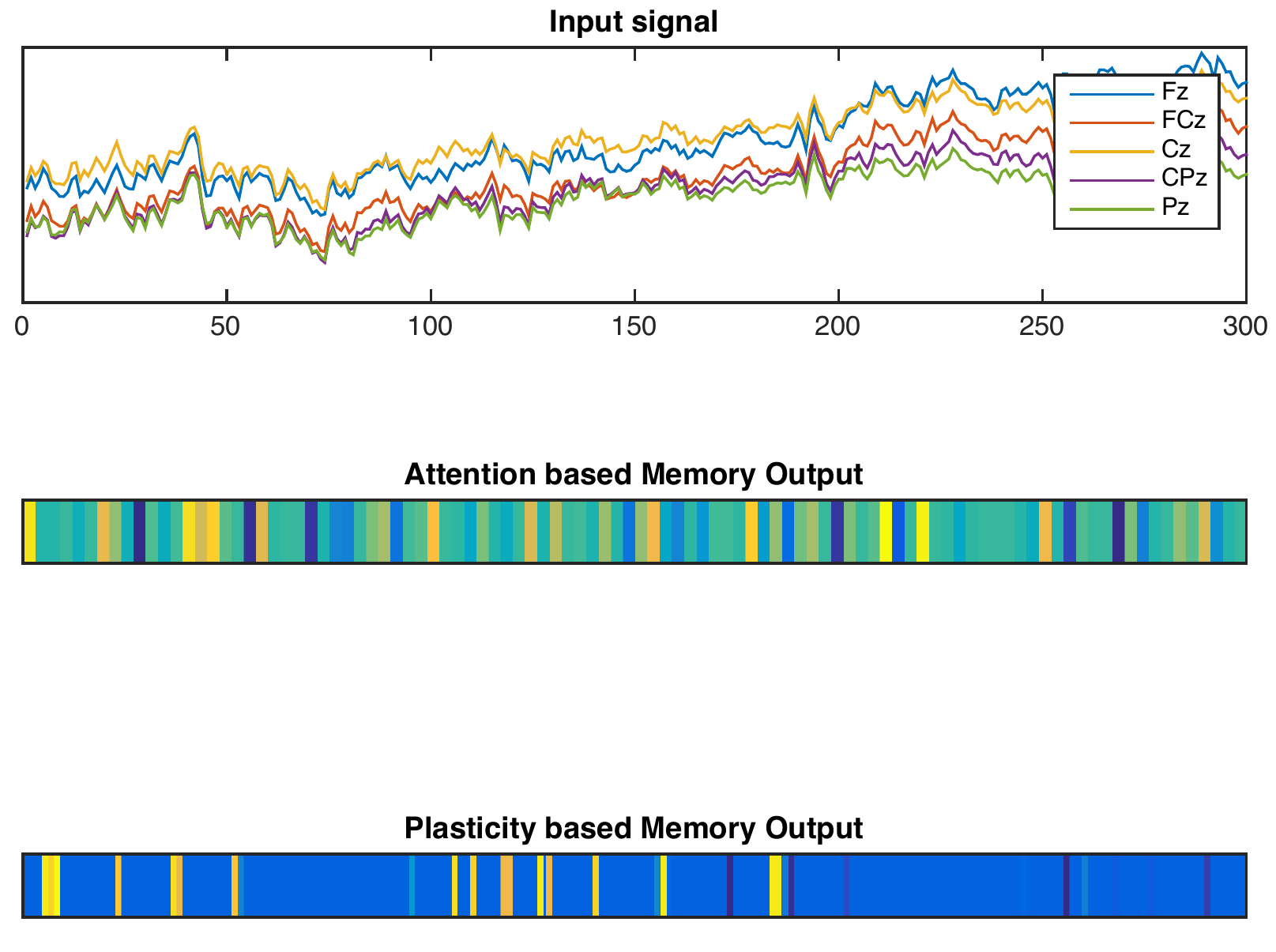}}
  \subfloat[][]{\includegraphics[width=.43\textwidth]{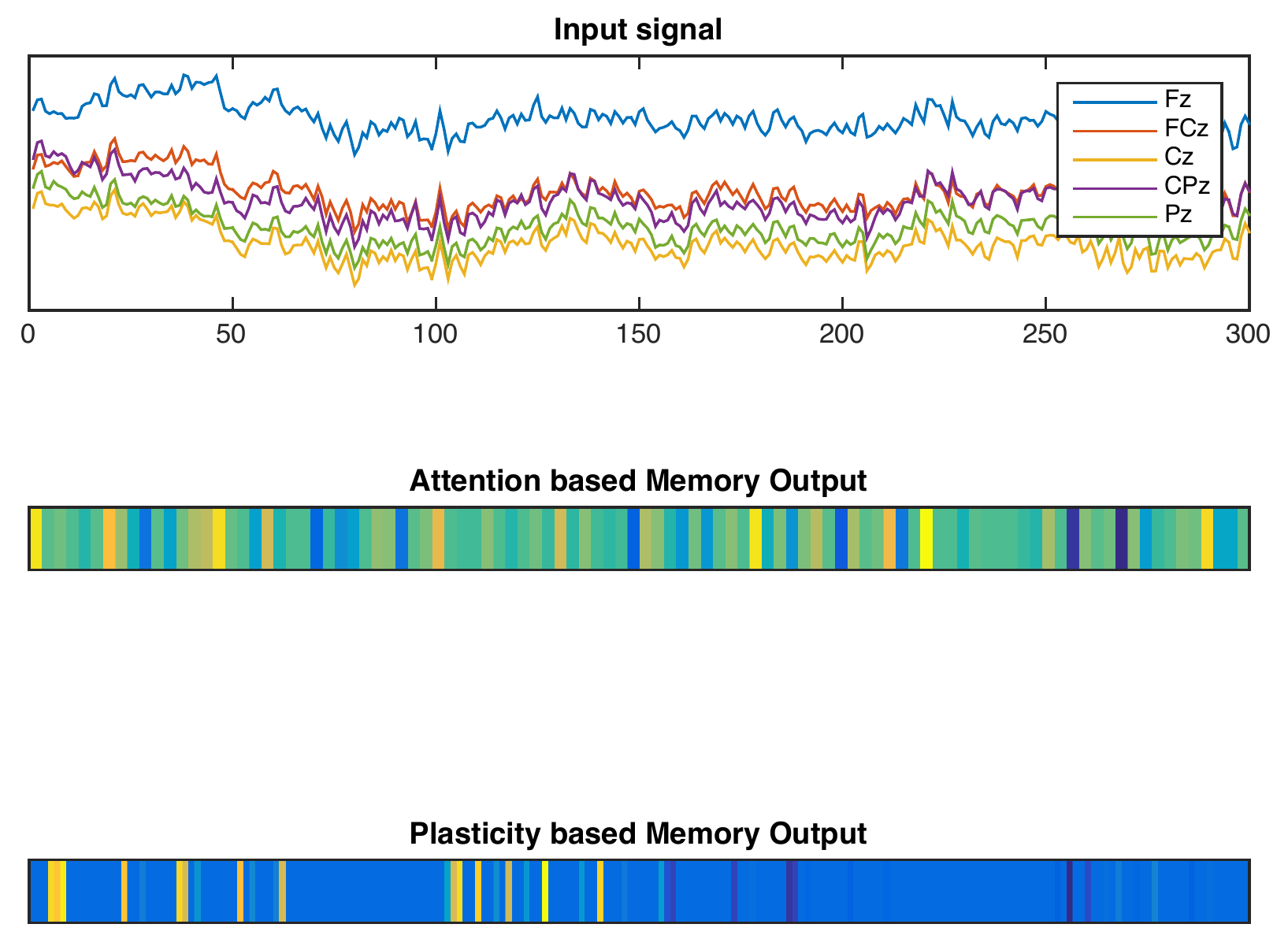}}
  \caption{Visualisation of memory outputs from the memory model of \cite{munkhdalai2017neural} (i.e Eq. \ref{eq:memory_out}) and the Plastic NMN (Proposed) method (i.e Eq. \ref{eq:plasticity_3}). Colours blue to yellow corresponds to low-high values.}
\label{fig:discuss_1}
\end{figure*}

In Fig. \ref{fig:discuss_1} we visualise the extracted activations for 3 sample inputs, where colours from blue to yellow correspond to low to high output values. From this illustration it is evident that the memory outputs are significantly sparser in the proposed method, which clearly exhibits that neural plasticity is able to systematically strengthen/ weaken the connections based on their importance, and this leads to the identification of the most salient information for decision making. 

\begin{figure}[htbp]
\centering
  \includegraphics[width=.75\linewidth]{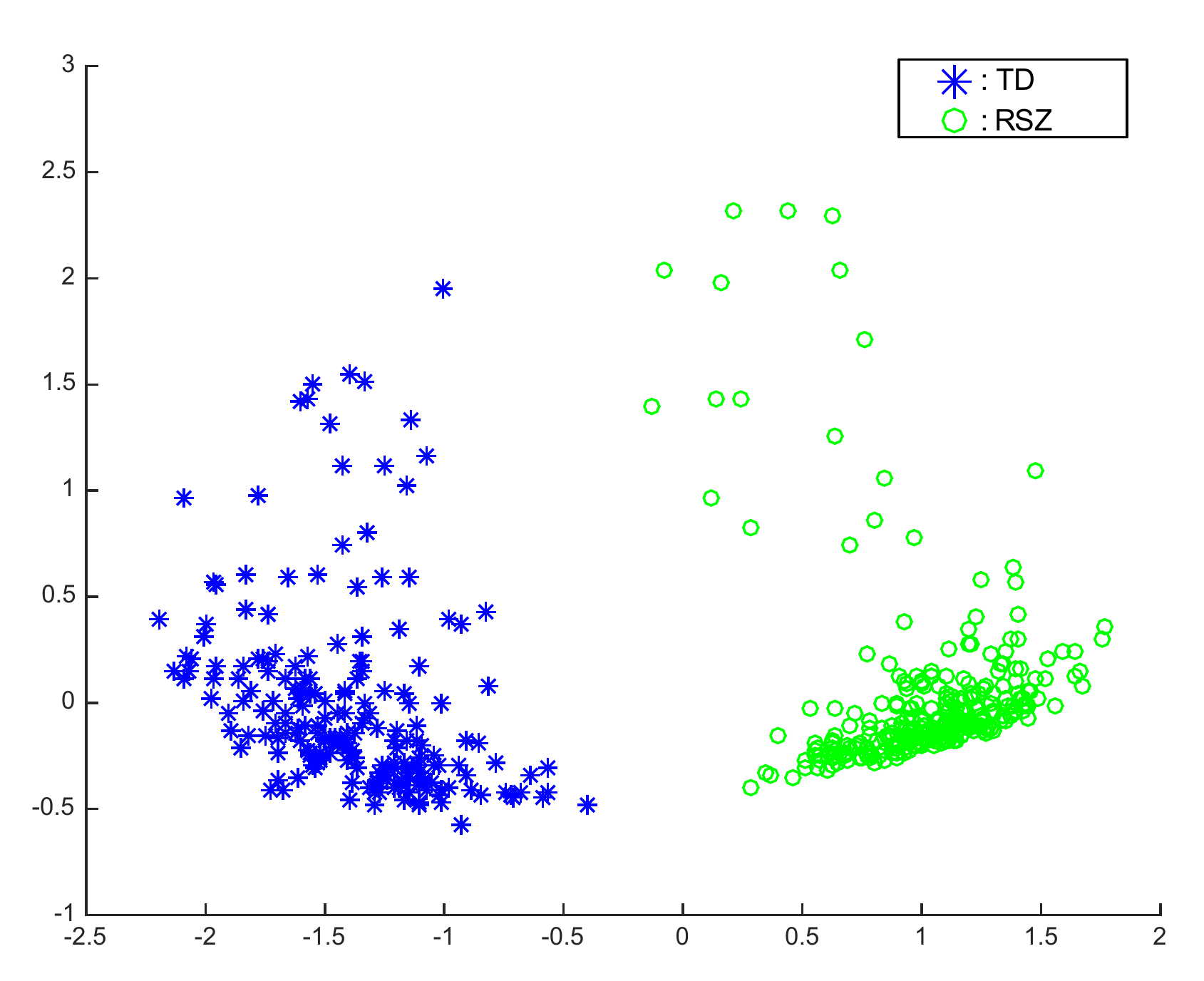}
  \caption{2D illustration of extracted memory embeddings for 500 randomly selected samples from the test set of the schizophrenia risk detection task presented in Sec. \ref{sec:eeg_schizophrenia}.}
\label{fig:discuss_5}
\end{figure}
\begin{figure}[htbp]
\centering
  \includegraphics[width=.75\linewidth]{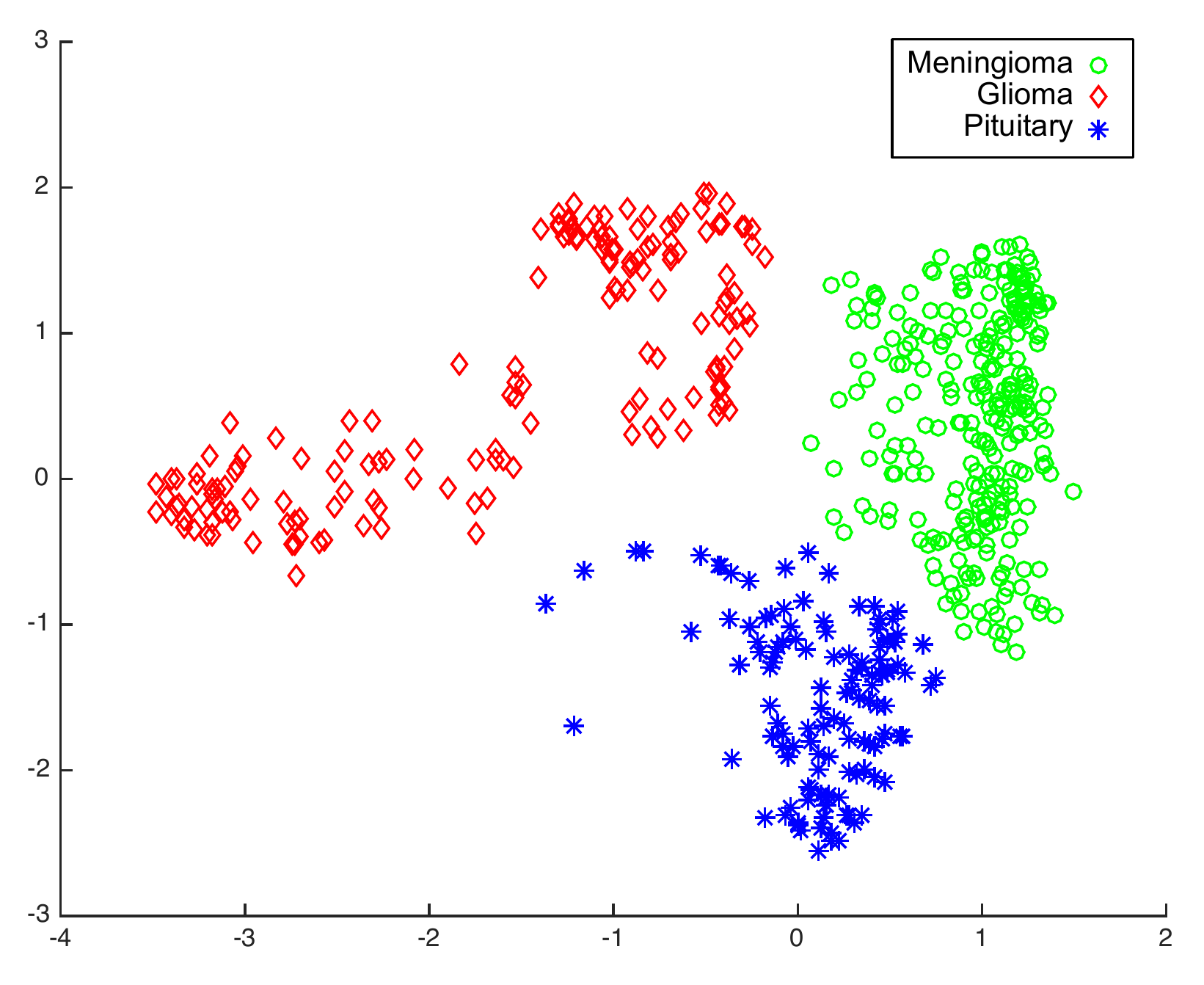}
  \caption{2D illustration of extracted memory embeddings for 500 randomly selected samples from the test set of the MRI tumour type classification task in Sec. \ref{sec:mri}.}
\label{fig:discuss_6}
\end{figure}

In addition, using Figs \ref{fig:discuss_5} and  \ref{fig:discuss_6} we provide memory embedding space visualisations in order to illustrate how the proposed memory network discriminates between different classes in the schizophrenia risk detection task presented in Sec. \ref{sec:eeg_schizophrenia} as well as the multi-class MRI tumour type classification problem presented in Sec. \ref{sec:mri}. 

We randomly sample 500 inputs from the test set and we apply PCA \cite{jolliffe2011principal} and plot these embeddings in 2D. Fig. \ref{fig:discuss_5} presents the resultant plot where the TD and RSz classes are indicated based on the ground truth class identities. We observe a clear separation between the TD and RSz classes. In Fig. \ref{fig:discuss_6} we illustrate the memory embedding plot for the MRI tumour type classification task in Sec. \ref{sec:mri}. Similar to the previous analysis we extract memory outputs for a randomly selected sample of 500 MRI inputs from the test set and applied PCA to plot the memory outputs in 2D. We have indicated glioma, meningioma and pituitary tumour types in red diamonds, blue starts and green circles, respectively based on their ground truth classes. We observe clear separation between the three classes. This clearly demonstrates that even though plasticity has led to additional sparsity in the memory output, the resultant sparse vectors are sufficient to discriminate between the classes. This highlights the deficiencies with the attention based memory output generation process, which leads to much denser outputs and incurs additional overhead on the classification layer as it has to discriminate between dense vectors. 

\begin{figure*}[htbp]
\centering
 \subfloat[][attention weights of baseline memory]{\includegraphics[width=.3\textwidth]{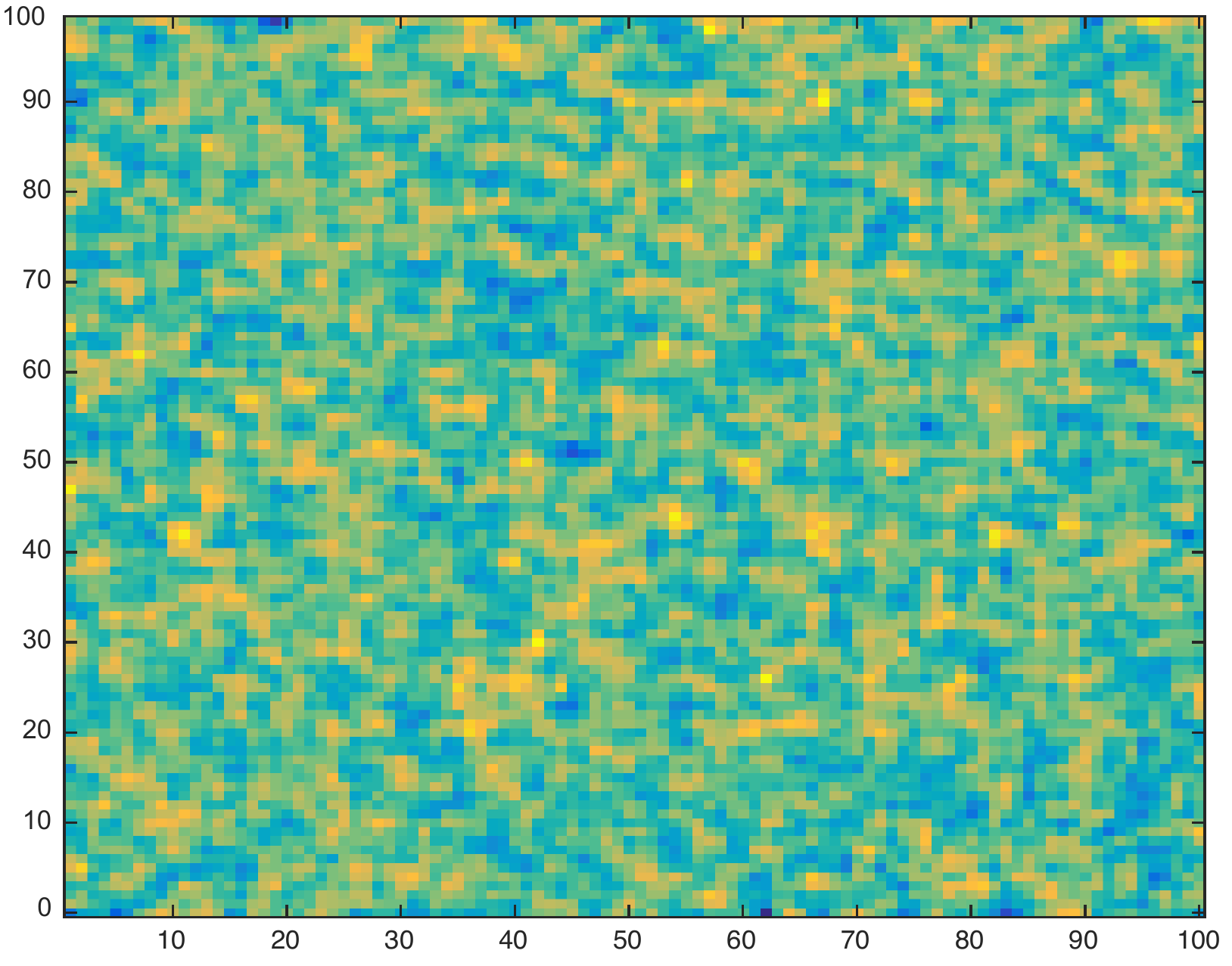}}
  \subfloat[][fixed weight, $\hat{w}$, of the output controller]{\includegraphics[width=.3\textwidth]{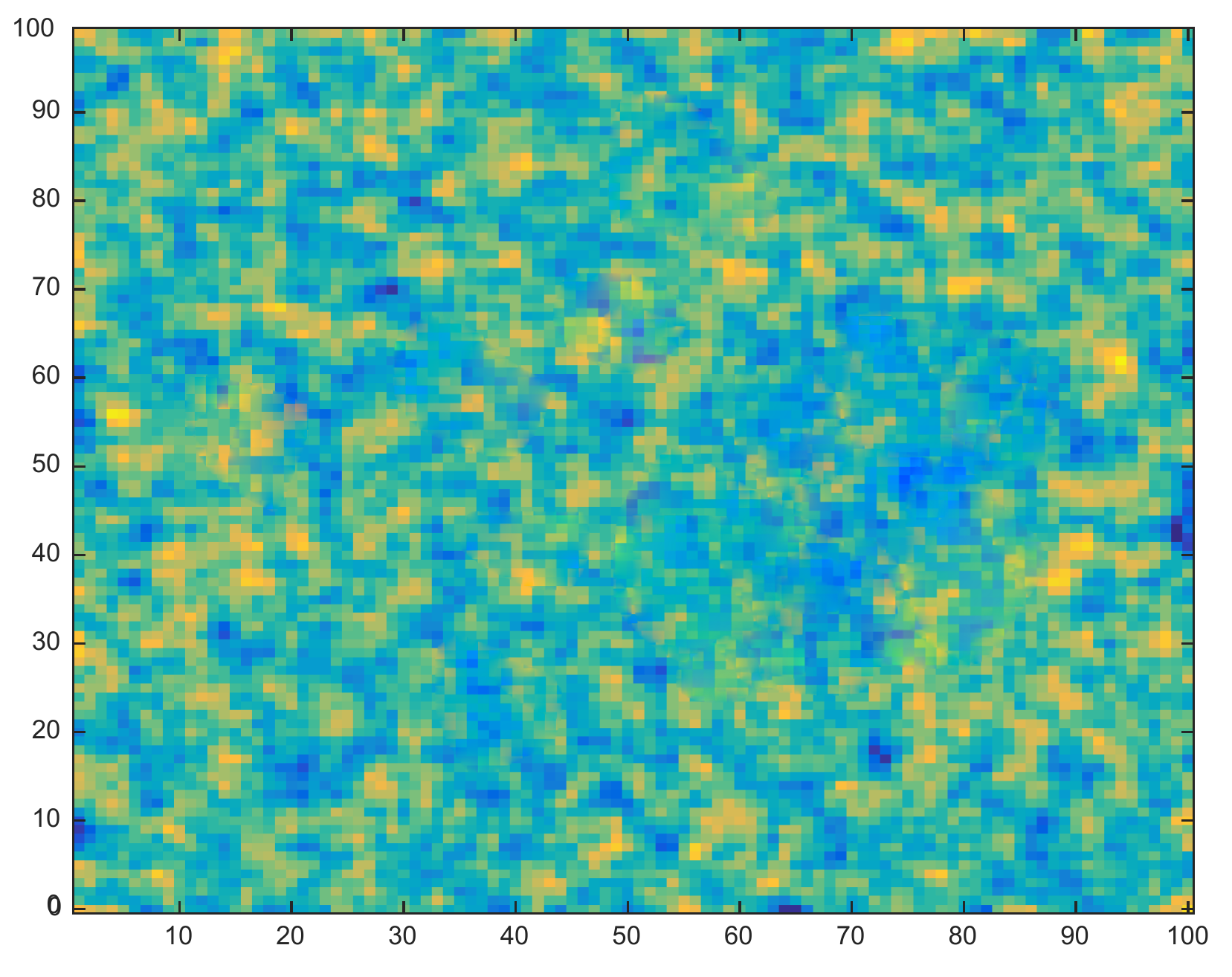}}
  \subfloat[][Hebbian trace, $\mathrm{\hat{Hebb}}$, of the output controller]{\includegraphics[width=.3\textwidth]{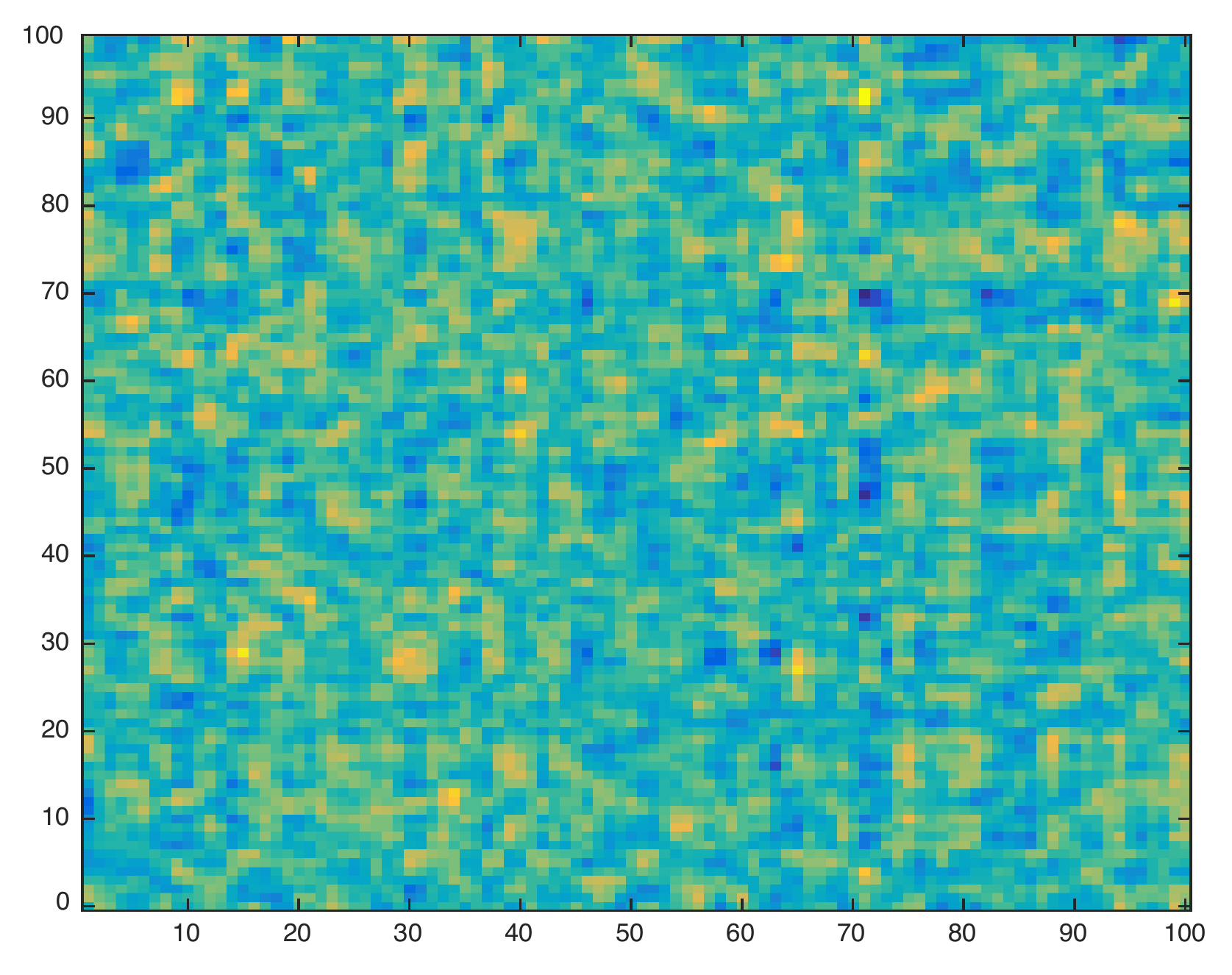}}
  \caption{Visualisation of attention weights of the baseline memory, and the fixed weight and Hebbian trace of the proposed output controller for the EEG Based Schizophrenia Detection Evaluation in Sec. \ref{sec:eeg_schizophrenia}. Colours blue to yellow correspond to low-high connection strengths.}
  \label{fig:discuss_2}
\end{figure*}

\begin{figure*}[htbp]
\centering
 \subfloat[][attention weights of baseline memory]{\includegraphics[width=.3\textwidth]{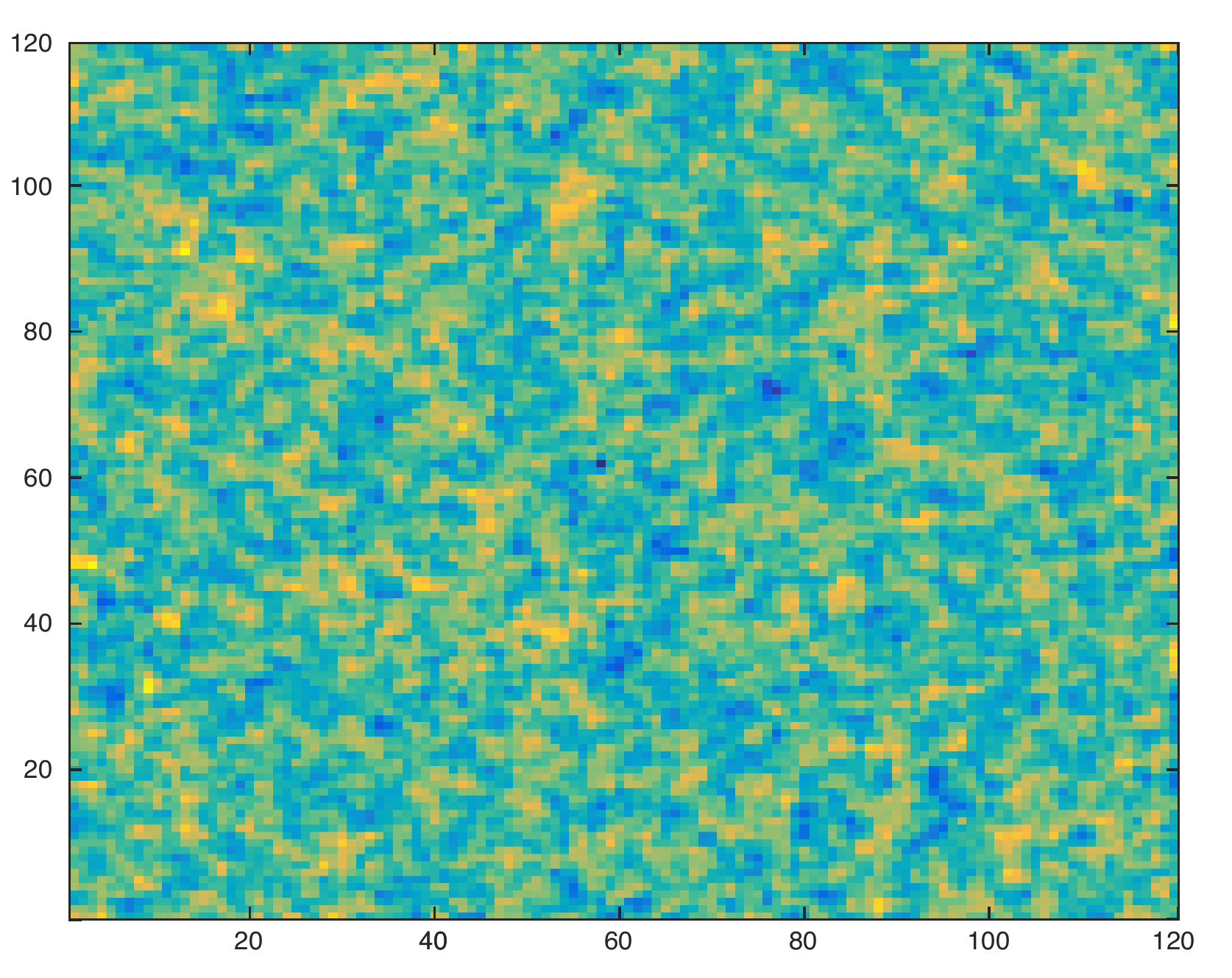}}
  \subfloat[][fixed weight, $\hat{w}$, of the output controller]{\includegraphics[width=.3\textwidth]{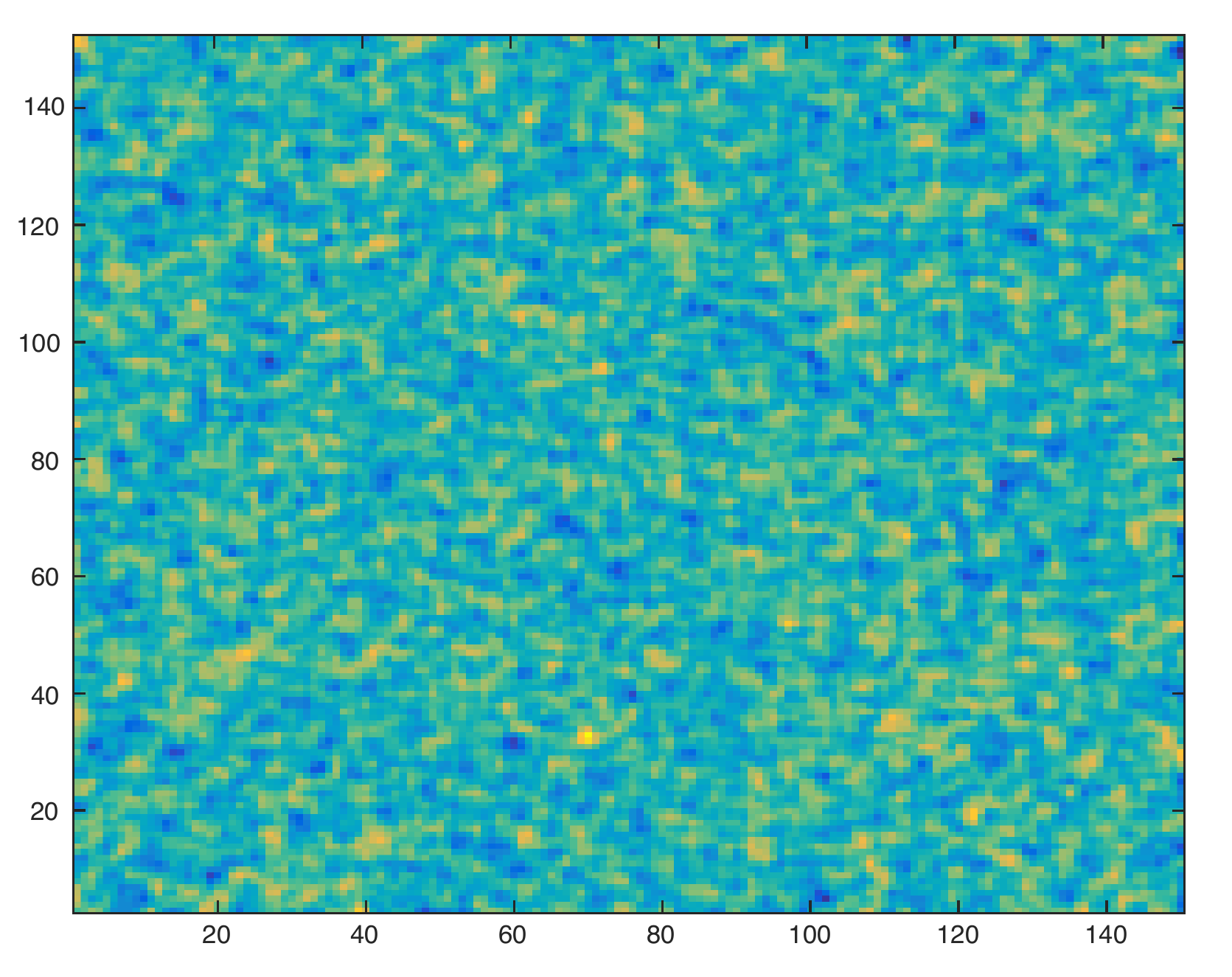}}
  \subfloat[][Hebbian trace, $\mathrm{\hat{Hebb}}$, of the output controller]{\includegraphics[width=.3\textwidth]{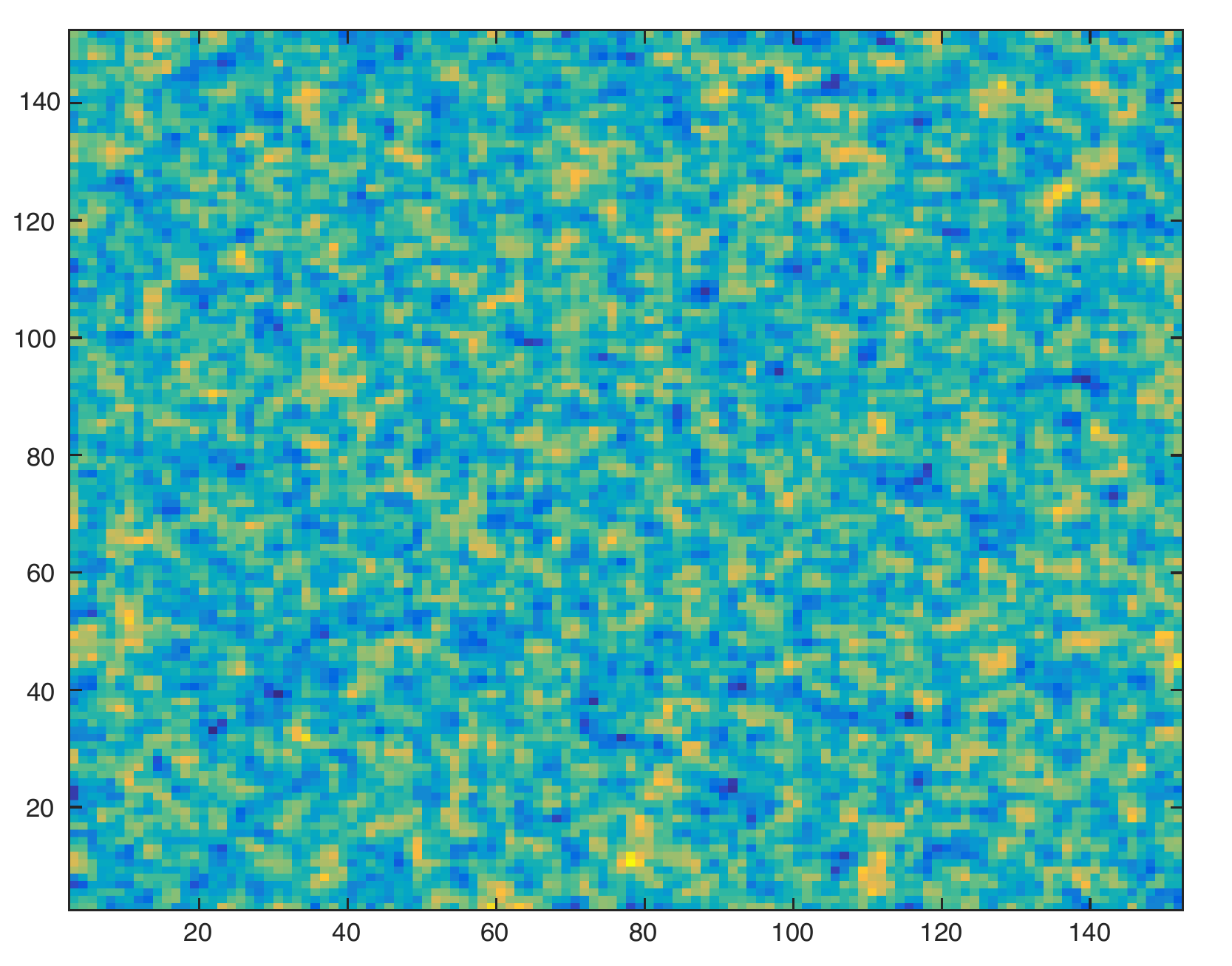}}
  \caption{Visualisation of attention weights of the baseline memory, and the fixed weight and Hebbian trace of the proposed output controller for the MRI Tumour Type Classification Evaluation in Sec. \ref{sec:mri}. Colours blue to yellow correspond to low-high connection strengths.}
\label{fig:discuss_2_2}
\end{figure*}

To further demonstrate the strengths of neural plasticity in the proposed method, in Figs. \ref{fig:discuss_2} and \ref{fig:discuss_2_2} we illustrate the attention weights of the baseline memory model of \cite{munkhdalai2017neural} , the fixed weights $\hat{w}$ and $\mathrm{\hat{Hebb}}$ of the output controller of the proposed method, once the training has completed, for the EEG based schizophrenia risk detection evaluation in Sec. \ref{sec:eeg_schizophrenia} and MRI tumour type classification evaluation in Sec. \ref{sec:mri}, respectively. In the EEG based schizophrenia risk detection evaluation as $k=100$ each of the plots are of dimension $100 \times 100$. In Fig. \ref{fig:discuss_2_2} we observe dimensions $150 \times 150$ as $k$ is set to 150 in the MRI tumour type classification evaluation for the proposed model and the attention weights of the baseline memory model has a dimension of $120 \times 120$. Colours from blue to yellow correspond to low-high connection weight values. 

When analysing the plot it is clear that the attention weight matrix of the baseline memory module is denser compared to the fixed weights in $\hat{w}$, demonstrating that only a subset of connections are required in all the scenarios. It should be emphasised that in the baseline memory model the attention weights are fixed once the training completes. In contrast, the $\mathrm{\hat{Hebb}}$ of the proposed method evolves over time and its visualisation illustrates a subset of connections that change over time to retrieve the salient components that are required at that particular time step. We argue that this process facilitates the sparser fixed weight matrix, as this can simply encode core information common to all patients, with the Hebb able to adapt to extract salient information for each individual case.

\begin{figure*}[htbp]
\centering
  \subfloat[][$\mathrm{\hat{Hebb}}$ after the training completes]{\includegraphics[width=.3\textwidth]{figures/hebbian/hebb_hat.pdf}}
  \subfloat[][$\mathrm{\hat{Hebb}}$ after the testing completes]{\includegraphics[width=.301\textwidth]{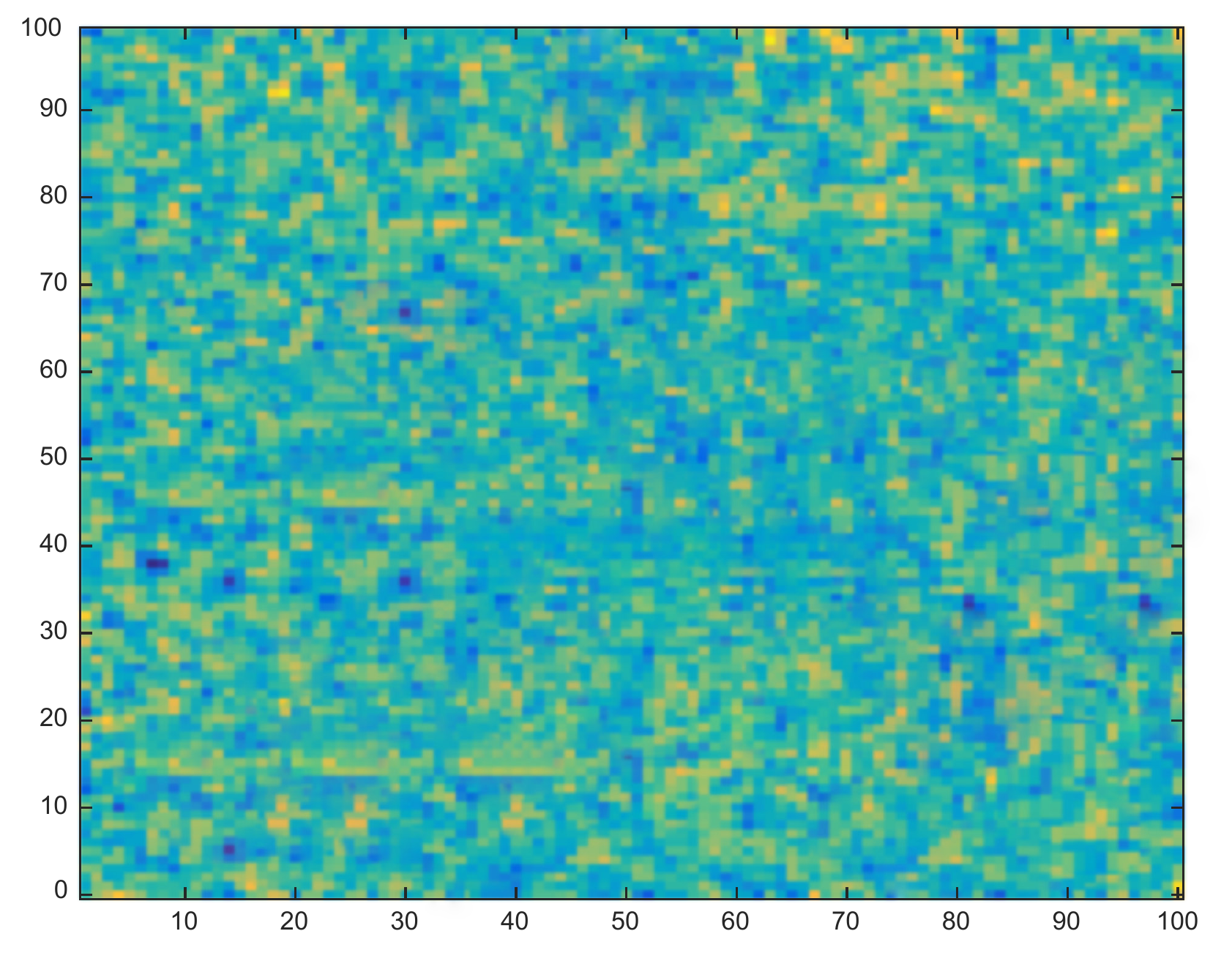}}
  \caption{Visualisation of the Hebbian trace, $\mathrm{\hat{Hebb}}$, evolving over time. }
\label{fig:discuss_2_3}
\end{figure*}

Fig. \ref{fig:discuss_2_3} visualises how the Hebbian trace, $\mathrm{\hat{Hebb}}$, changes between the training and testing instances. We visualise the $\mathrm{\hat{Hebb}}$ once the training has completed (i.e. at the start of testing) and when the testing process has been completed (i.e. once it has seen all samples in the testing dataset). The changes in the Hebbian trace clearly demonstrates that different connections are needed in order to extract salient information in different cases, as opposed to having fixed connections throughout. 

\subsection{What is the model learning?}

In this subsection we attempt to interpret what the proposed model detects as salient when distinguishing between RSz and TD groups in the experiment outlined in Sec. \ref{sec:eeg_schizophrenia}. 

\begin{figure*}[htbp]
\centering
 \subfloat[][]{\includegraphics[width=.43\textwidth,height=3cm]{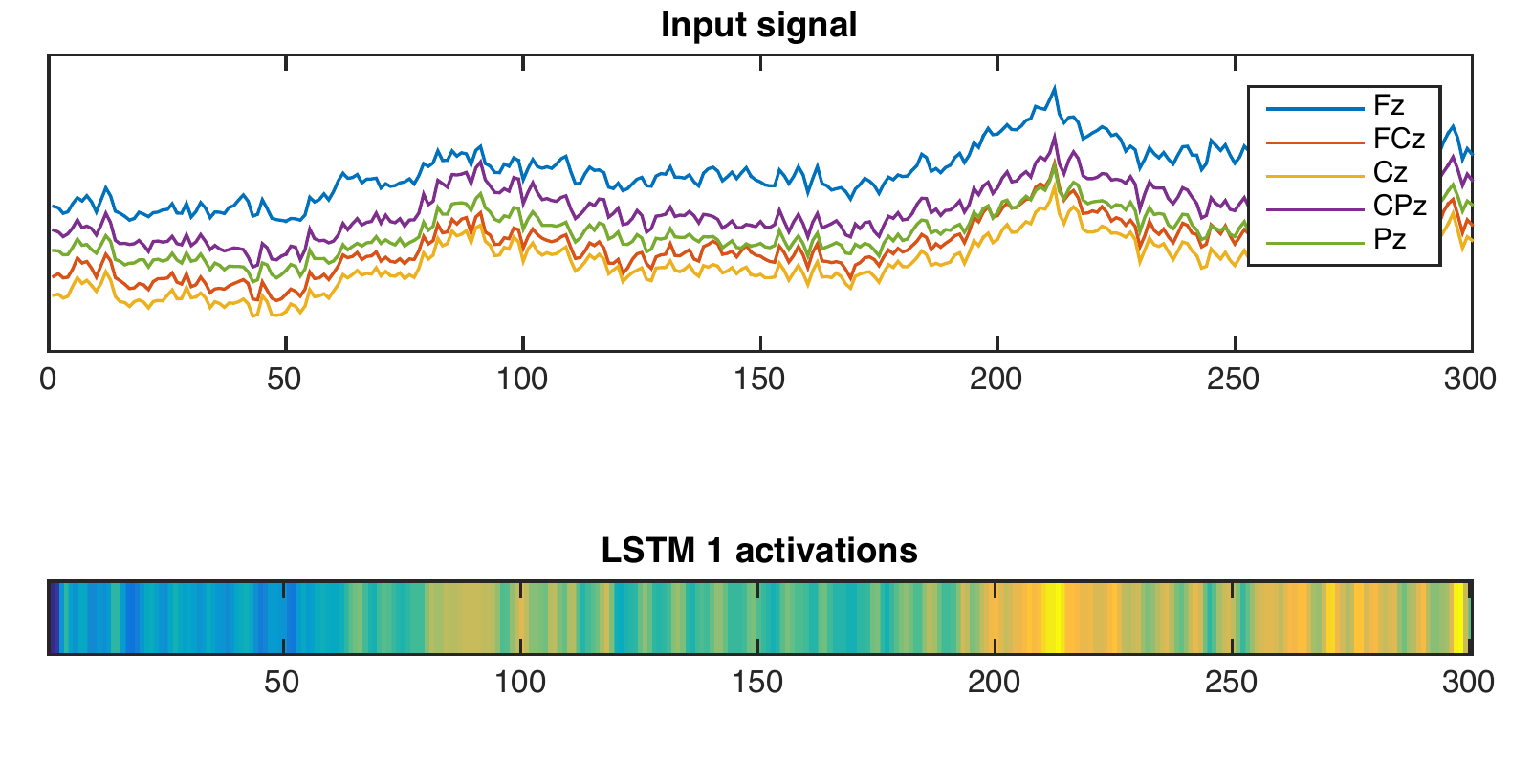}}
 \subfloat[][]{\includegraphics[width=.43\textwidth,height=3.07cm]{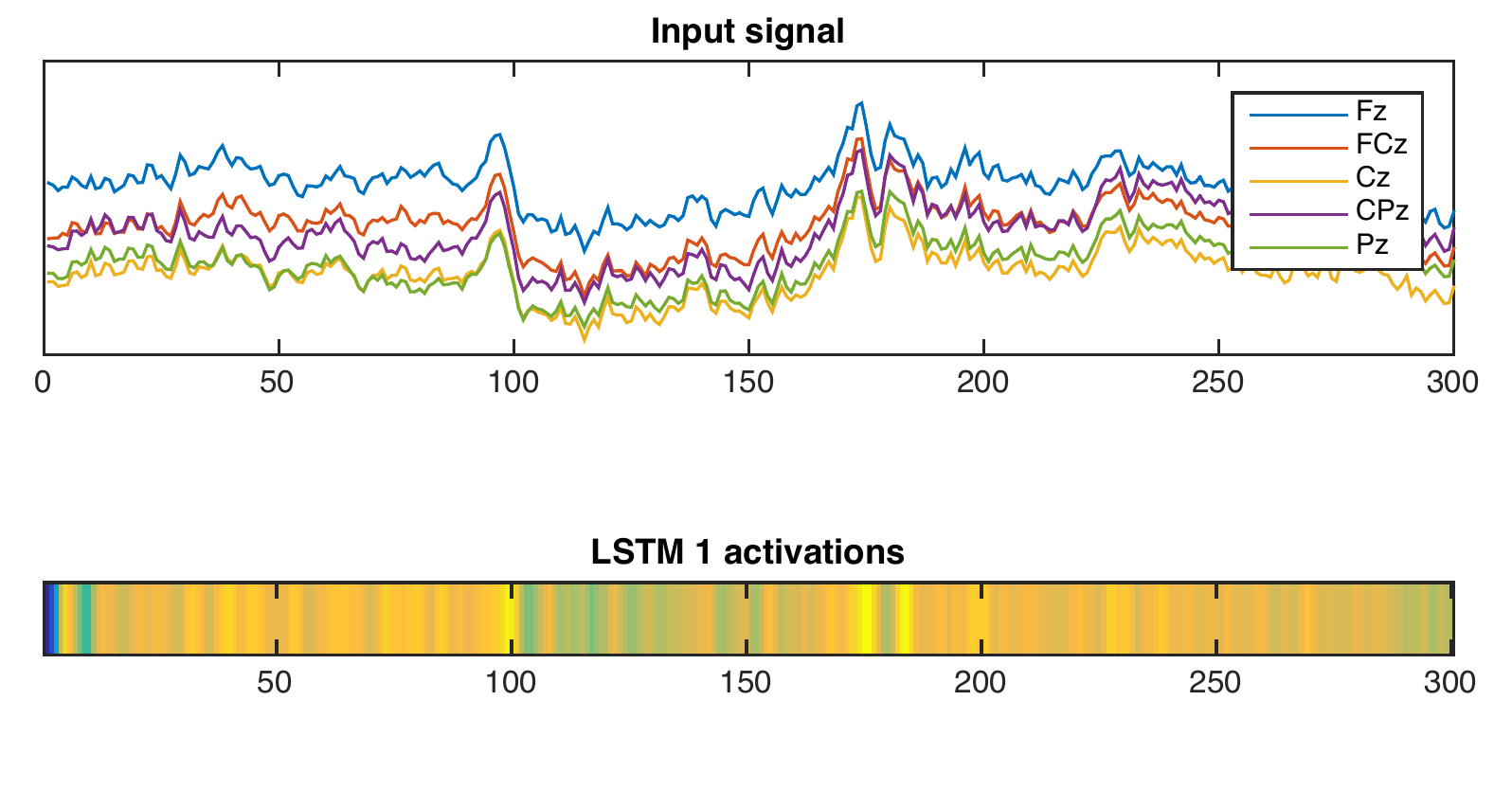}} \\
 \subfloat[][]{\includegraphics[width=.43\textwidth,height=3cm]{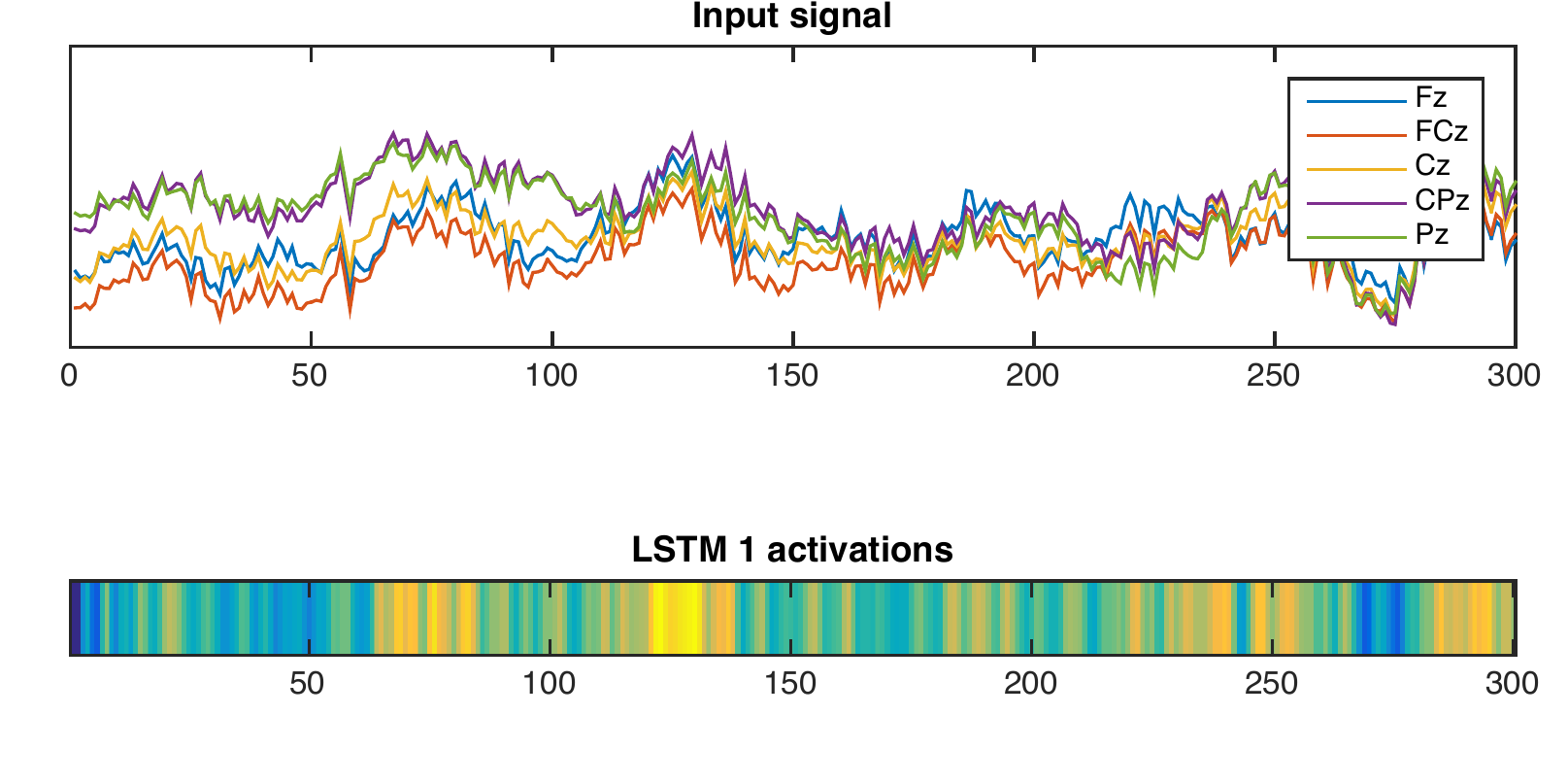}}
  \subfloat[][]{\includegraphics[width=.43\textwidth,height=3cm]{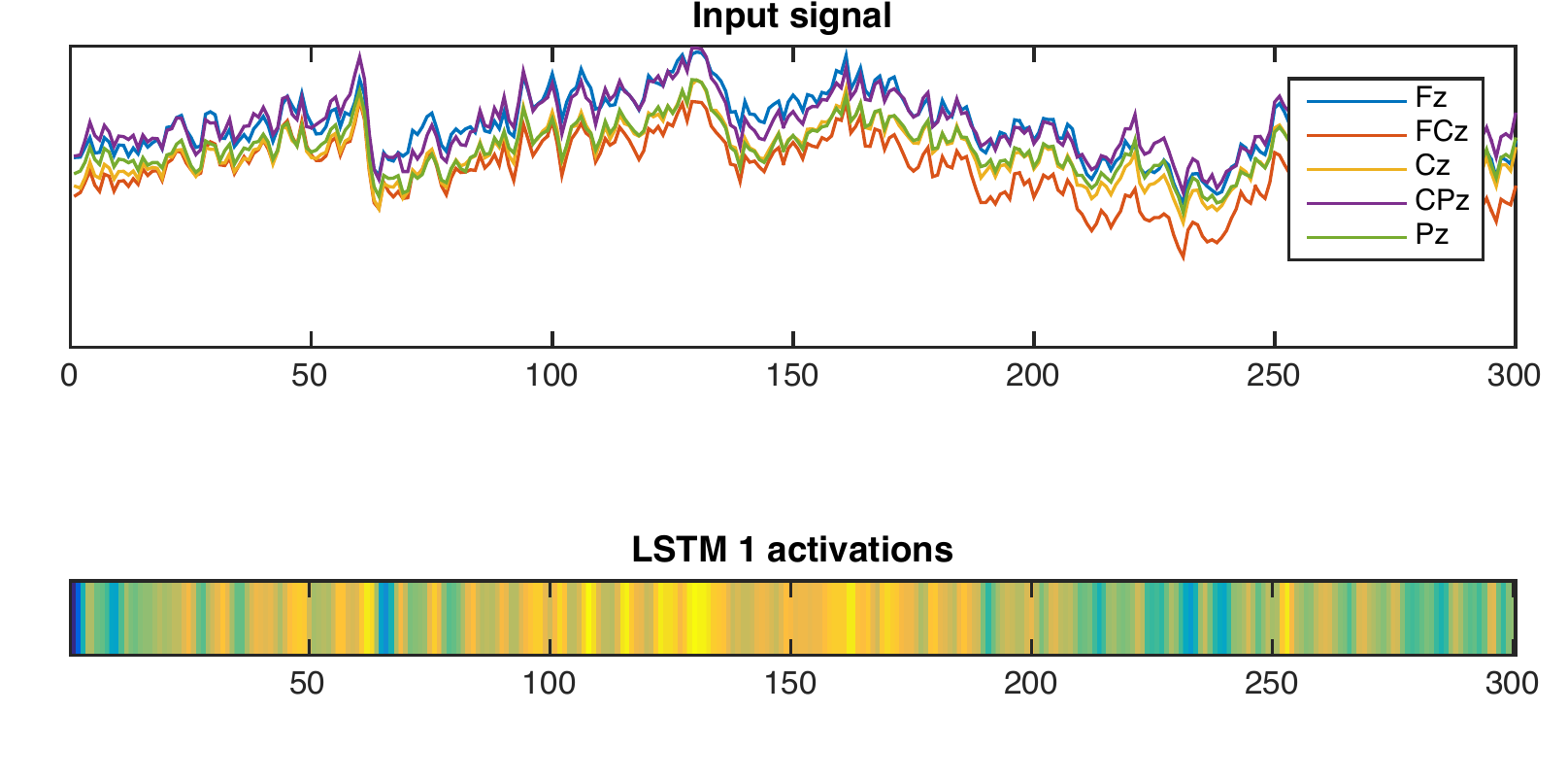}}
  \caption{Activations from the first LSTM layer of the proposed model and its corresponding inputs.}
\label{fig:discuss_3}
\end{figure*}

First, we extract activations from the first LSTM layer of the proposed model (See Fig. \ref{fig:full_model}) for four randomly selected examples which are presented in Fig. \ref{fig:discuss_3}. The model provides 
more attention to the areas of the input EEG which corresponds to sudden fluctuations in the waveforms, identifying important events such as peaks and valleys generated in the mismatch trials.

\begin{figure}[htbp]
\centering
  \includegraphics[width=.75\linewidth]{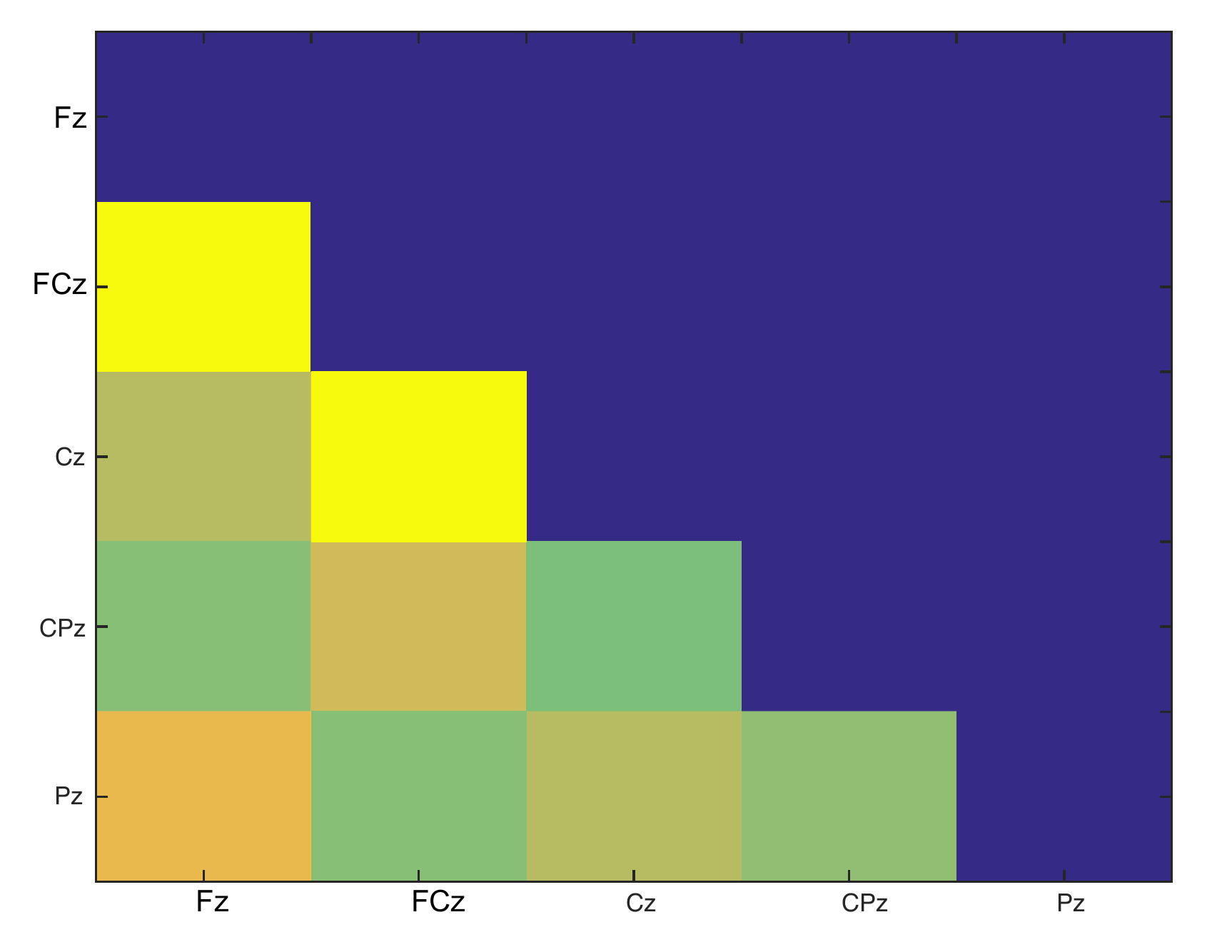}
  \caption{Visualisation of feature importance of EEG channels in terms of the pariwise interactions based analysis proposed in \cite{tsang2017detecting}. Colours blue to yellow corresponds to low-high importance.}
\label{fig:discuss_4}
\end{figure}

In order to determine the exact channels that correspond to decision making we adapt the statistical interaction analysis framework proposed in \cite{tsang2017detecting}. This method ranks the neural network weights of the input layer on its statistical interactions that are performed with its first hidden layer. In particular the utilised pairwise interaction ranking scheme ranks all pairs of input features according to their interaction strengths. Fig. \ref{fig:discuss_4} illustrates the output. As our input contains 5 input channels (i.e Fz, FCz, Cz, CPz and Pz) the figure illustrates the possible pairwise connections between these 5 inputs. We observe higher interaction strengths between channels Fz - FCz and FCz - Cz which supports the findings reported in \cite{bruggemann2013mismatch} which also indicates the higher degree of activities in brain frontal lobe in the auditory mismatch paradigm. 

\subsection{Hardware and Time Complexity Details}
The implementation of the proposed model is completed using Keras \cite{chollet2015keras} with a Theano \cite{bergstra2010theano} backend. The proposed model does not require any special hardware such as GPUs to run and the model used for the abnormal EEG identification task presented in Sec. \ref{sec:abnormal_EEG_Detection} has 170K trainable parameters. We measured the time complexity of the proposed method using the test set of THU EEG database \cite{lopez2015automated} used in Sec. \ref{sec:abnormal_EEG_Detection}. The proposed model is capable of generating 1000 predictions ( i.e using 1000 input sequences of 60 second in lengths and generates 1000 classifications) in 13.16 seconds on a single core of an Intel Xeon E5-2680 2.50 GHz CPU.

\begin{figure}
\centering
  \subfloat[][]{\includegraphics[width=.48\linewidth]{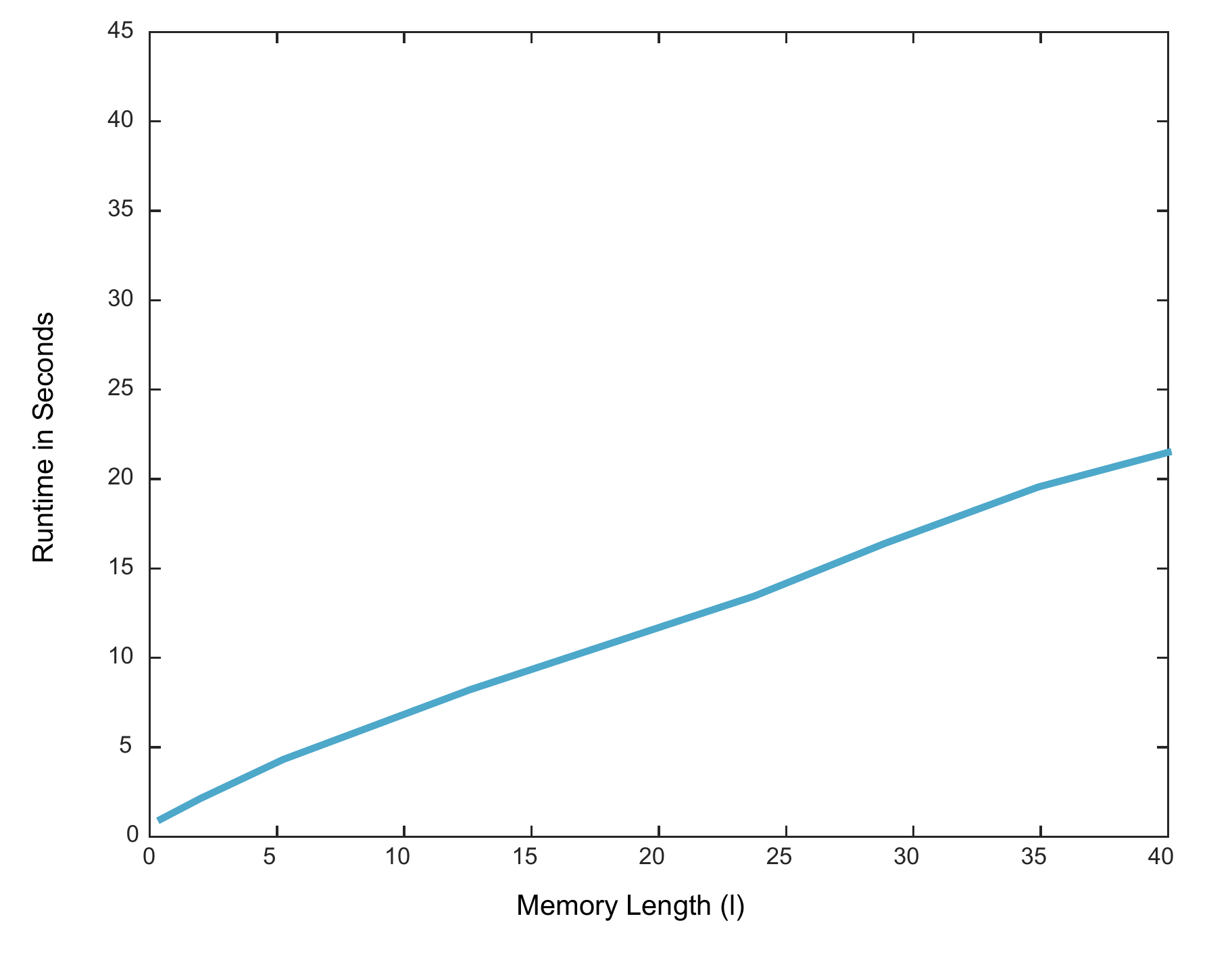}}
 \subfloat[][]{\includegraphics[width=.45\linewidth]{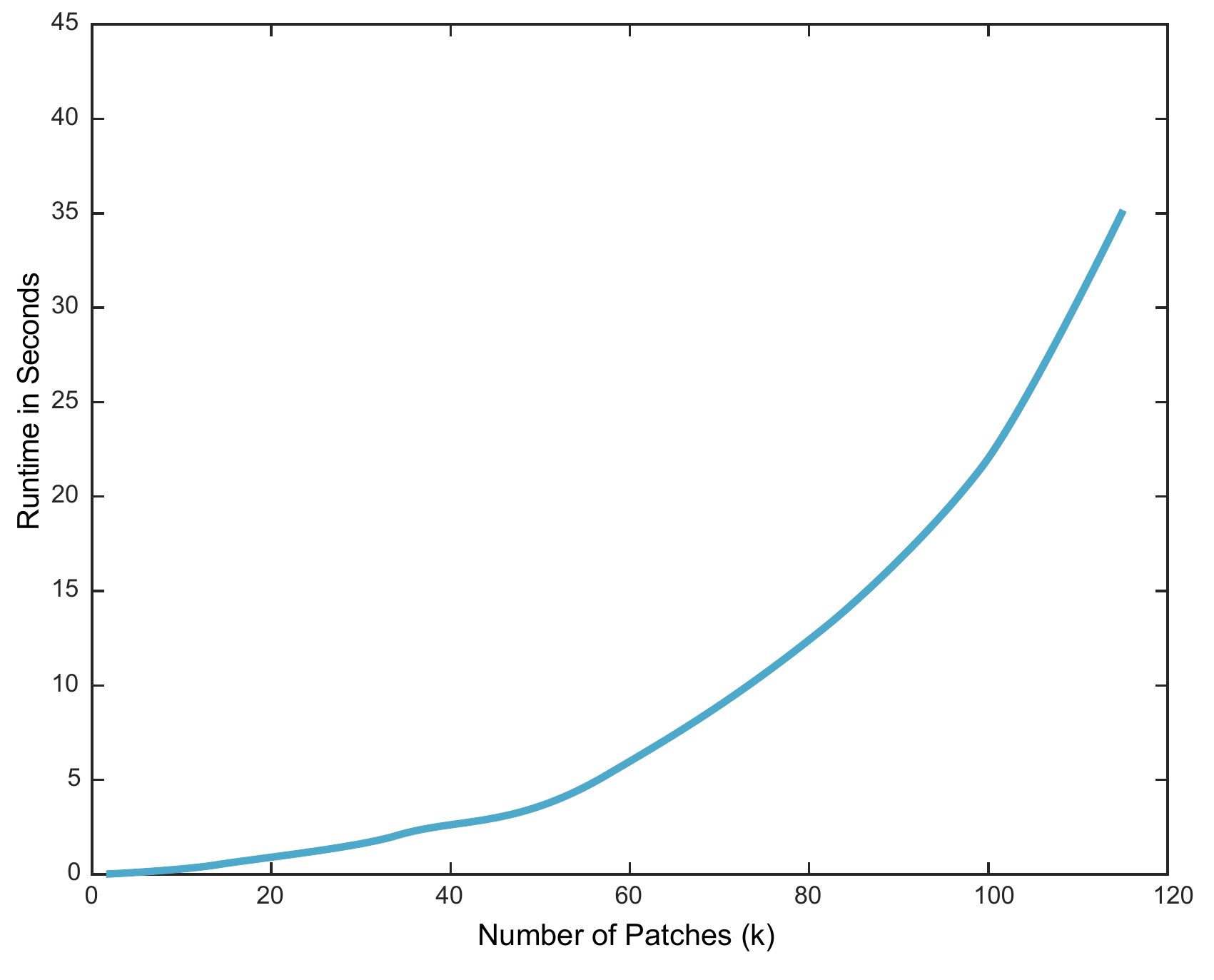}}
\caption{Evaluation of runtimes for different memory length, $l$ (a), and embedding dimensions, $k$, values (b).}
\label{fig:runtimes}
\end{figure}

In the same experimental setting, we measured the time required to generate 1000 predictions for different lengths of the memory module, $l$, and different embedding dimensions, $k$. Results are given in Fig. \ref{fig:runtimes}. With the memory length, $l$, the runtime grows approximately linearly, while with the embedding dimension, $k$, it grows exponentially. This is because an increase in embedding dimension increases the dimensionalities of the fixed weight and plasticity components in the memory which incur significant additional computational overhead. 

\section{Conclusion}
In this paper, we propose a plastic neural memory network architecture which exploits the the advances in Neural Memory Networks (NMNs) and neural plasticity. We introduce plasticity in memory access mechanism which allows the underlying framework to pay varying level of attention to different problem specific and subject specific information that it requires to retrieve from the stored knowledge within the memory. We point out and illustrate the drawbacks of current attention-based knowledge retrieval processes in NMNs and demonstrate how the neural plasticity can be used to overcome these deficiencies. Through the evaluation conducted on three challenging anomaly detection tasks in the medical domain, we demonstrate that our proposed memory architecture is able to outperform all considered baselines. Through visualisation of the of the mechanisms of the proposed memory architecture, we provide evidence of the power of our our memory addressing process to capture salient information cues that are needed for anomaly detection. The varied nature of the evaluations, which includes both one and two-dimensional data, demonstrates how the proposed model can be directly applied to any anomaly detection or classification problem where modelling long term relationships is necessary. In future work, we will be exploring the applications of neural memory plasticity for encoding multimodal inputs and where the sparsity of the generated memory outputs can be utilised to summarise and represent denser input representations. 

\section*{Acknowledgment}
K.R.L was supported by an Australian Research Council Future Fellowship (FT170100294). Funding for collection of EEG data in the schizophrenia risk sample was provided by a National Institute for Health Research (UK) Career Development Fellowship (CDF/08/01/015) and BIAL Foundation Research Grants (36/06 and 194/12).


\bibliography{egdb}

\end{document}